\documentclass[journal]{journal}
\usepackage{amsmath,amsfonts,bm}

\def\eqref#1{equation~\ref{#1}}
\def\1{\bm{1}}

\DeclareMathAlphabet{\mathsfit}{\encodingdefault}{\sfdefault}{m}{sl}
\SetMathAlphabet{\mathsfit}{bold}{\encodingdefault}{\sfdefault}{bx}{n}

\def\gA{{\mathcal{A}}}

\def\gM{{\mathcal{M}}}
\def\gN{{\mathcal{N}}}

\def\gS{{\mathcal{S}}}
\def\gT{{\mathcal{T}}}

\def\pitheta{\pi_{\theta}}
\def\vpsi{V_{\psi}}
\def\rphi{r_{\phi}}
\def\taue{\tau_E}
\def\pie{\pi_E}

\newcommand{\pzero}{p_{\rm{0}}}

\newcommand{\E}{\mathbb{E}}

\newcommand{\R}{\mathbb{R}}

\DeclareMathOperator*{\argmax}{arg\,max}

\usepackage{microtype}
\usepackage{graphicx}
\usepackage[square,numbers]{natbib}
\usepackage[ruled,vlined,commentsnumbered,linesnumbered]{algorithm2e}
\usepackage{booktabs}       % professional-quality tables
\usepackage{amsfonts}       % blackboard math symbols
\usepackage{subcaption, graphicx}

\newcommand{\ie}{\textit{i.e.}}
\newcommand{\eg}{\textit{e.g.}}

\begin{document}
%
% paper title
% can use linebreaks \\ within to get better formatting as desired
\title{Imitating, Fast and Slow:
Robust learning from demonstrations via decision-time planning}
%
%
% author names and IEEE memberships
% note positions of commas and nonbreaking spaces ( ~ ) LaTeX will not break
% a structure at a ~ so this keeps an author's name from being broken across
% two lines.
% use \thanks{} to gain access to the first footnote area
% a separate \thanks must be used for each paragraph as LaTeX2e's \thanks
% was not built to handle multiple paragraphs
%

% \author{Michael~Shell,~\IEEEmembership{Member,~IEEE,}
%         John~Doe,~\IEEEmembership{Fellow,~OSA,}
%         and~Jane~Doe,~\IEEEmembership{Life~Fellow,~IEEE}% <-this % stops a space
% \thanks{M. Shell is with the Department
% of Electrical and Computer Engineering, Georgia Institute of Technology, Atlanta,
% GA, 30332 USA e-mail: (see http://www.michaelshell.org/contact.html).}% <-this % stops a space
% \thanks{J. Doe and J. Doe are with Anonymous University.}% <-this % stops a space
% \thanks{Manuscript received April 19, 2005; revised January 11, 2007.}}

\author {
    % Authors
    Carl Qi,\textsuperscript{\rm 1}
    Pieter Abbeel, \textsuperscript{\rm 2}
    Aditya Grover \textsuperscript{\rm 3} \\
    \textsuperscript{\rm 1} Carnegie Mellon University \\
    \textsuperscript{\rm 2} University of California, Berkeley \\
    \textsuperscript{\rm 2} Facebook AI Research \\
}

% The only time the second header will appear is for the odd numbered pages
% after the title page when using the twoside option.
% 
% *** Note that you probably will NOT want to include the author's ***
% *** name in the headers of peer review papers.                   ***
% You can use \ifCLASSOPTIONpeerreview for conditional compilation here if
% you desire.

\maketitle

\begin{abstract}
    
The goal of imitation learning is to mimic expert behavior from demonstrations, without access to an explicit reward signal.
A popular class of approach infers the (unknown) reward function via inverse reinforcement learning (IRL) followed by maximizing this reward function via reinforcement learning (RL).
% to learn an imitation policy.
The policies learned via these approaches are however very brittle in practice and deteriorate quickly even with small test-time perturbations due to compounding errors.
We propose \textit{Imitation with Planning at Test-time} (IMPLANT), a meta-algorithm for imitation learning that utilizes decision-time planning to correct for compounding errors of any base imitation policy.
In contrast to existing approaches, we retain both the imitation policy and the rewards model at decision-time, thereby benefiting from the learning signal of the two components.
% disentangling and mutually countering imperfections due to reward inference (via IRL) and policy optimization (via RL).
Empirically, we demonstrate that IMPLANT significantly outperforms benchmark imitation learning approaches on standard control environments and excels at zero-shot generalization when subject to challenging perturbations in test-time dynamics.
\end{abstract}

\begin{IEEEkeywords}
Deep Learning, Imitation Learning, Inverse Reinforcement Learning, Zero-shot Generalization
\end{IEEEkeywords}

\section{Introduction}

The objective of imitation learning is to optimize agent policies directly from demonstrations of expert behavior.
Such a learning paradigm sidesteps reward engineering, which is a key bottleneck for applying reinforcement learning (RL) in many real-world domains from robotics to autonomous driving.
With only a finite dataset of expert demonstrations, however, a key challenge is that the learned policies can quickly deviate from intended expert behavior and lead to compounding errors at test-time~\cite{osa2018algorithmic}.
Moreover, it has been observed that imitation policies can be brittle and drastically deteriorate in performance with even small perturbations to the dynamics during execution~\citep{christiano2016transfer,de2019causal}. 

A predominant class of approaches to imitation learning is based on inverse reinforcement learning (IRL) and involve successive application of two steps: (a) an IRL step where the agent infers a proxy for the (unknown) reward function for the expert, followed by (b) an RL step where the agent maximizes the inferred reward function via a policy optimization algorithm.
For example, many popular IRL approaches consider an adversarial learning framework~\citep{goodfellow2014generative}, where the reward function is inferred by a discriminator that distinguishes expert demonstrations from rollouts of an imitation policy [IRL step] and the imitation agent maximizes the inferred reward function to best match the expert policy [RL step]~\citep{ho2016generative,fu2017learning}.
In this sense, reward inference is only an intermediary step towards learning the expert policy and is discarded post-training of the imitation agent.

We introduce \textit{Imitation with Planning at Test-time} (IMPLANT), a new meta-algorithm for imitation learning that incorporates decision-time planning into an IRL algorithm. 
During training, we can use any standard IRL approach to estimate a reward function and a stochastic imitation policy, along with an additional value function.
The value function can be learned explicitly or is often a byproduct of standard RL algorithms that involve policy evaluation, such as actor-critic methods~\citep{konda2000actor,peters2008natural}.
At decision-time, we use the learned imitation policy in conjunction with a closed-loop planner.
For any given state, the imitation policy proposes a set of candidate actions and the planner estimates the returns for each of actions by performing fixed-horizon rollouts.
The rollout returns are estimated using the learned reward and value functions.
Finally, the agent picks the action with the highest return and the process is repeated at each of the subsequent timesteps.
\begin{figure*}[t]
    \centering
    \begin{subfigure}[b]{\linewidth}
        \centering
        \begin{subfigure}[b]{0.32\linewidth}
        \centering
        \includegraphics[width=\linewidth]{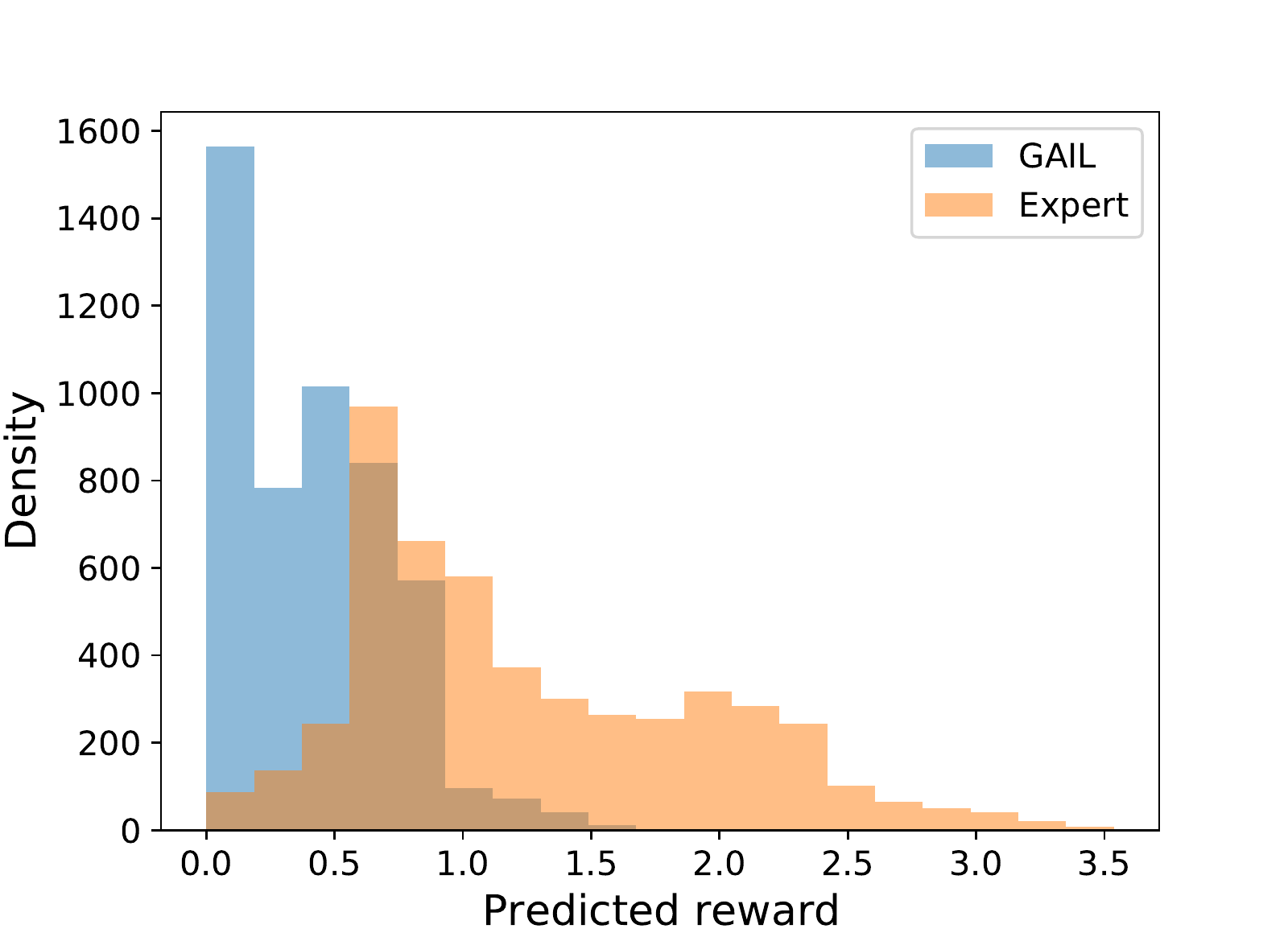}
        \caption{Hopper}
        \end{subfigure}
        \centering
        \begin{subfigure}[b]{0.32\linewidth}
        \centering
        \includegraphics[width=\linewidth]{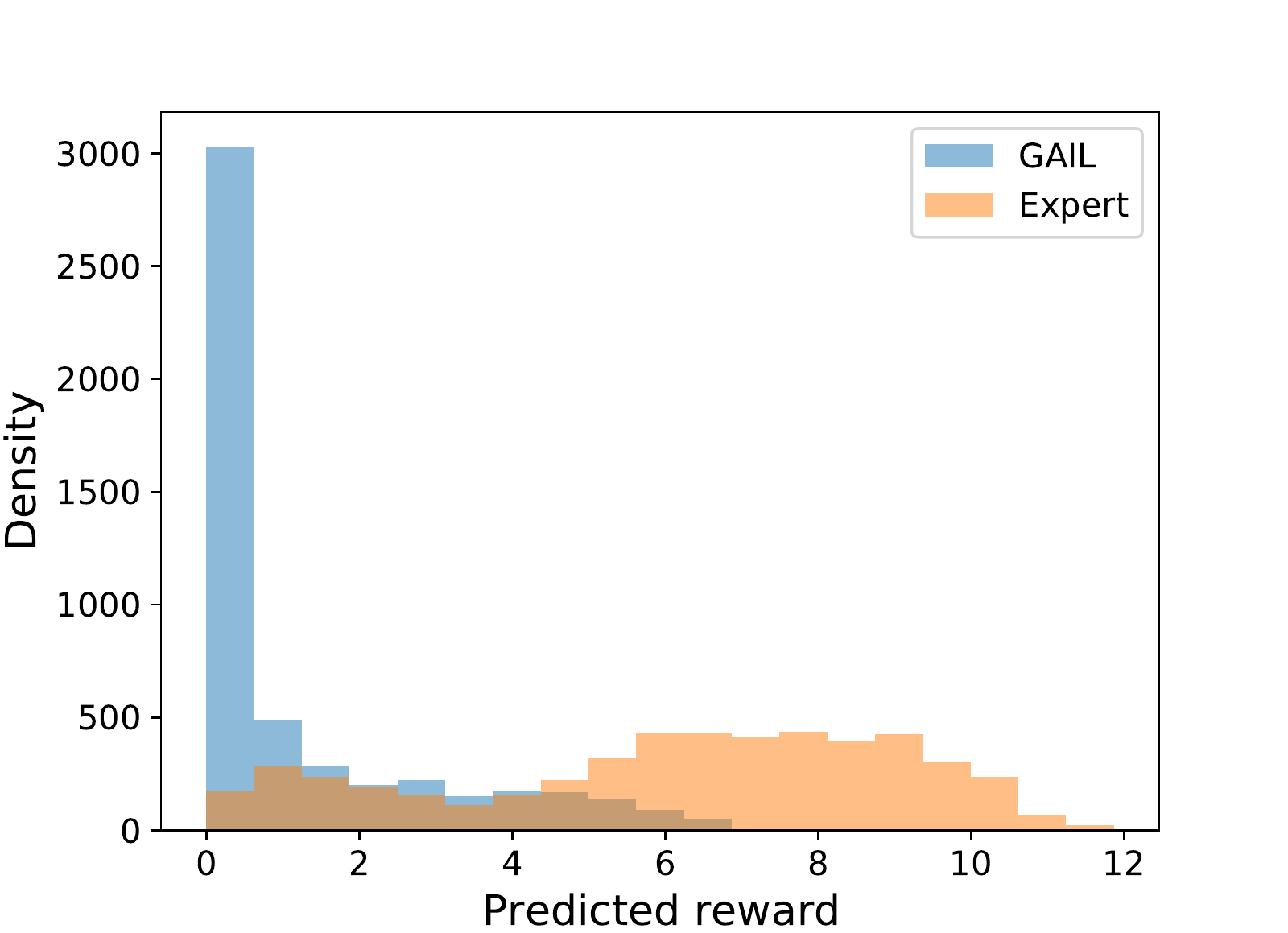}
        \caption{HalfCheetah}
        \end{subfigure}
        \begin{subfigure}[b]{0.32\linewidth}
        \centering
        \includegraphics[width=\linewidth]{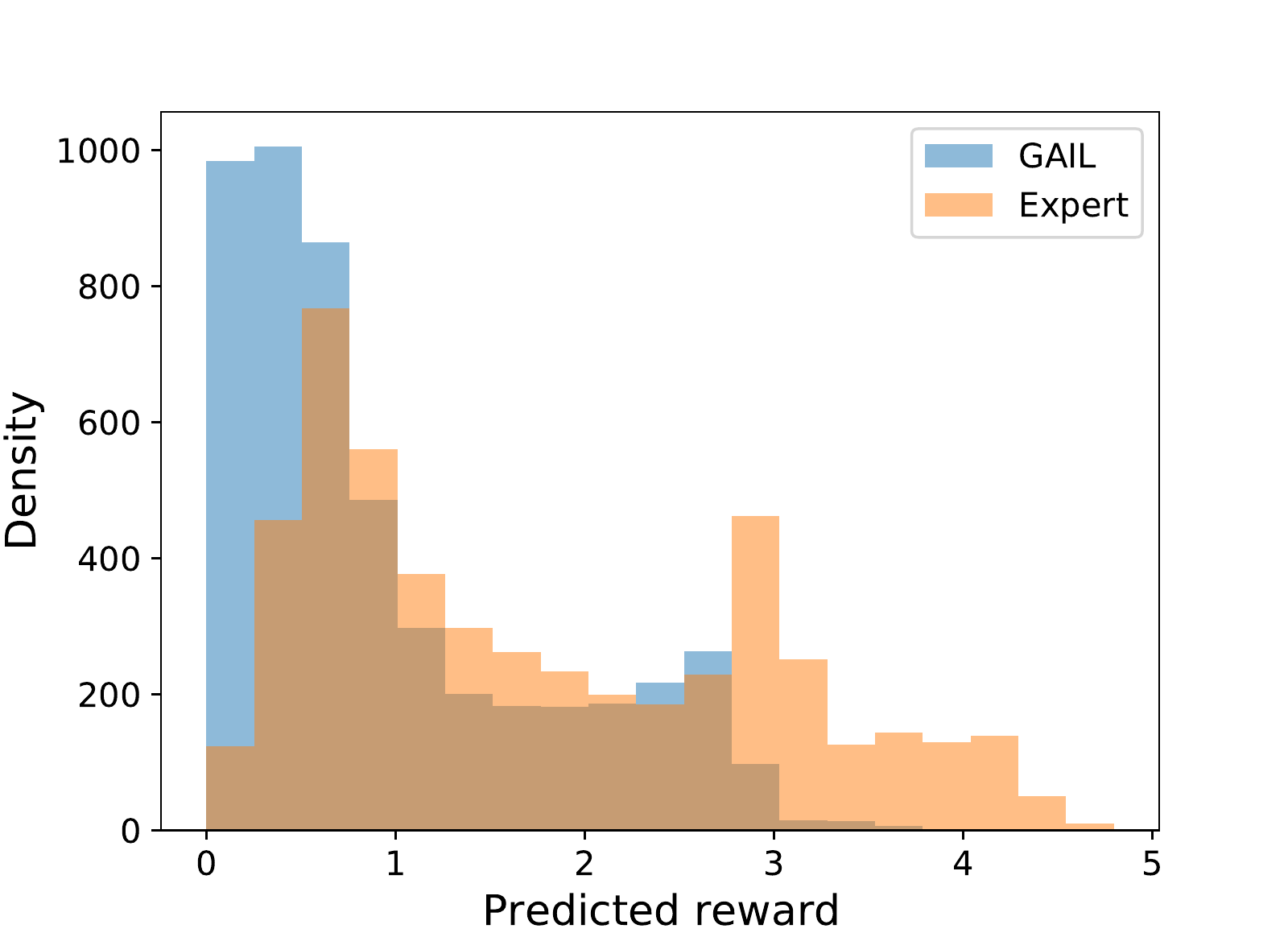}
        \caption{Walker2d}
        \end{subfigure}
    \end{subfigure}
    \caption{Distribution of predicted rewards for trained GAIL policies vs. experts.}\label{fig:gail_disc}
\end{figure*}

The use of decision-time planning can slow the reaction time of the agent, but can offer significant performance gains.
Conceptually, IMPLANT aims to counteract the imperfections due to policy optimization in the RL step by using the proxy reward function (along with a value function) estimated in the IRL step for decision-time planning.
We demonstrate strong empirical improvements using this approach over benchmark imitation learning algorithms in a variety of settings derived from the MuJoCo-based benchmarks in OpenAI Gym~\citep{todorov2012mujoco,brockman2016openai}.
In the default evaluation setup where train and test environments match, we observe that IMPLANT improves by 15.7\% on average over the closest baseline.

We also consider zero-shot transfer setups where the learned agent is deployed in test dynamics that differ from train dynamics and are inaccessible to the agent during both training and decision-time planning.
In particular, we consider the following setups: (a) ``causal confusion" where the agent observes nuisance variables in the state representation~\citep{de2019causal}, (b) motor noise which adds noise in the executed action~\citep{christiano2016transfer}, and (c) transition noise which adds noise to the sampled next state.
In all setups, we observe that IMPLANT consistently and robustly transfers to test environments with improvements of 52.1\% on average over the closest baseline.
% in policy returns.

\section{Background and Setup}

% \subsection{Problem Setup}
% \paragraph{Problem Setup.}
We consider the framework of Markov Decision Processes (MDP)~\citep{puterman1990markov}.
An MDP is denoted by a tuple $\gM = (\gS, \gA, \gT, \pzero, r, \gamma)$, where $\gS$ is the state space, $\gA$ is the action space, $\gT: \gS \times \gA \times \gS \to \R_{\geq 0}$ are the stochastic transition dynamics, $\pzero: \gS \to \R_{\geq 0}$ is the initial state distribution, $r: \gS \times \gA \to \R$ is the reward function, and $\gamma \in [0, 1)$ is the discount factor. 
We assume an infinite horizon setting.
At any given state $s \in \gS$, an agent makes decisions via a stochastic policy $\pi: \gS \times \gA \to \R_{\geq 0}$.
We denote a trajectory to be a sequence of state-action pairs $\tau=(s_0, a_0, s_1, a_1, \cdots)$.
Any policy $\pi$, along with MDP parameters, induces a distribution over trajectories, which can be expressed as $p_\pi(\tau)= p(s_0) \prod_{t=0}^{\infty} \pi(a_t \vert s_t) \gT(s_{t+1} \vert s_t, a_t)$.
% where $T>0$ is the horizon.
The return of a trajectory is the discounted sum of rewards $R(\tau) = \sum_{t=0}^{\infty} \gamma^t r(s_t, a_t)$.

In reinforcement learning (RL), the goal is to learn a parameterized policy $\pi_\theta$ that maximizes the expected returns w.r.t. the trajectory distribution. 
% Formally, we minimize the following loss function:
% \begin{align}\label{eq:rl}
%   \ell_{RL}(\theta) := -\E_{\tau\sim p_{\pi_\theta}} \left[R(\tau)\right] .
% \end{align}
% Many practical algorithms and variants exist, such as MaxEntRL which adds an entropy regularization term to the standard RL objective~\citep{kappen2005path}. 
Maximizing such an objective requires interaction with the underlying MDP for simulating trajectories and querying rewards. 
However, in many high-stakes scenarios, the reward function is not directly accessible and hard to manually design.
% We denote a stochastic policy as $\pi: \gS \times \gA \to \R$.

In imitation learning, we sidestep the availability of the reward function.
% is unknown and inaccessible to the agent.
Instead, we have access to a finite set of $D$ trajectories $\taue$ (a.k.a. demonstrations) that are sampled from an expert policy $\pie$. 
% Typically, we hope that the expert policy to be close to the optimal policy $\pi^\ast$ w.r.t. the RL objective in Eq.~\ref{eq:rl}.
Every trajectory $\tau \in \taue$ consists of a finite-length sequence of state and action pairs $\tau = (s_0, a_0, s_1, a_1, \cdots)$, where $s_0 \sim \pzero(s)$, $a_t \sim \pie( \cdot \vert s_t)$, and $s_{t+1} \sim \gT(\cdot \vert s_t, a_t)$.
Our goal is to learn a parameterized policy $\pi_\theta$ which best approximates the expert policy given access to $\taue$.
Next, we discuss the two major families of techniques for imitation learning.
% Throughout this work, we assume that all parametric functions are specified via deep neural networks.

\subsection{Behavioral Cloning}\label{sec:bc}

Behavioral cloning (BC) casts imitation learning as a supervised learning problem over state-action pairs provided in the expert demonstrations~\citep{pomerleau1991efficient}.
In particular, we learn the policy by solving a regression problem with states $s_t$ and actions $a_t$ as the features and target labels respectively.
% The states $s_t$ serve as features and the corresponding actions $a_t$ are the target labels for regression.
Formally, we minimize the following objective:
\begin{align}
   \ell_{BC}(\theta) := \sum_{(s_t, a_t) \in \taue } \Vert a_t - \pi_\theta(s_t) \Vert_2^2.
\end{align}

In practice, BC agents suffer from \textit{distribution shift} in high dimensions, where small deviations in the learned policy accumulate during deployment and lead to a significantly different trajectory distribution relative to the expert~\citep{ross2010efficient,ross2011reduction}.
% Hence, in practice, BC agents .

\subsection{Inverse Reinforcement Learning}\label{sec:irl}
An alternative indirect approach to imitation learning is based on inverse reinforcement learning (IRL).
Here, the goal is to infer a reward function for the expert and subsequently maximize the inferred reward to obtain a policy.
For brevity, we focus on adversarial imitation learning approaches to IRL~\citep{goodfellow2014generative}.
These approaches represent the state-of-the-art in imitation learning and are also relevant baselines for our empirical evaluations.

Generative Adversarial Imitation Learning (GAIL) is an IRL algorithm that formulates imitation learning as an ``occupancy measure matching" objective w.r.t. a suitable probabilistic divergence~\citep{ho2016generative}.
GAIL consists of two parameterized networks: (a) a policy network $\pi_\theta$ (generator) which is used to rollout agent trajectories (assuming access to transition dynamics), and (b) a discriminator $D_\phi$ which distinguishes between ``real" expert demonstrations and ``fake" agent trajectories. 
Given expert trajectories $\taue$ and agent trajectories $\tau_\theta$, the discriminator minimizes the cross-entropy loss:
\begin{align}
    \ell_{IRL}(\phi) := - \E_{\taue} \left[\log D_\phi(\taue) \right] - \E_{\tau_\theta} \left[\log (1-D_\phi(\tau_\theta)) \right].
\end{align}

We then feed the discriminator output  $-\log(1-D_\phi(s, a))$ as the inferred reward function to the generator policy. The policy parameters $\theta$ can be updated via any standard policy optimization algorithm,
% in Eq.~\ref{eq:rl} or a related variant, 
\eg, \citet{ho2016generative} use the TRPO algorithm~\citep{schulman2015trust}.
% The objective function of the generator is as follows:
By simulating agent rollouts, GAIL seeks to match the full trajectory state-action distribution of the imitation agent with the expert as opposed to BC which greedily matches the conditional distribution of individual actions given the states. 
In practice, GAIL and its variants~\citep{li2017infogail,fu2017learning} outperform BC but might need excessive interactions with the training environment for sampling rollouts. 
Crucially, both BC and IRL approaches tend to fail catastrophically in the presence of small perturbations and nuisances at test-time~\citep{de2019causal}.

\section{The IMPLANT Framework}
% old values
% \begin{table*}
%     \centering
%     \caption{Average return of imitation learning algorithms on MuJoCo benchmarks.
%     % Number of expert demonstrations in parenthesis.
%     }
%     \vspace{0.5em}
%     \small
%     \begin{tabular}{lccc}
%     \toprule

%      & Hopper & HalfCheetah  & Walker2d  \\
%     \midrule
%     Expert              & $3570$        & $891$         & $3593$ \\
%     BC                  & $127 \pm 85 $ & $427 \pm 131$ & $258 \pm 262$ \\
%     BC-Dropout          & $169 \pm 105$ & $542 \pm 275$ & $1622 \pm 861$ \\
%     GAIL                & $3506 \pm 337$& $954 \pm 282$ & $2780 \pm 1007$ \\
%     % GAIL-Dropout        & $127 \pm 50$     & $-1 \pm 1$     & $269 \pm 6$ \\
%     GAIL-Reward Only      & $319 \pm 123$     & $5 \pm 117$     & $56 \pm 146$ \\
%     IMPLANT (ours)      & $\textbf{3633} \pm 50$ & $\textbf{1193} \pm 143$& $\textbf{3360} \pm 442$ \\
%     \bottomrule
%     \end{tabular}

%     \label{tab:1}
% \end{table*}

\begin{algorithm}[t]
    \SetAlgoLined
    % \SetKw{KwInput}{\textbf{Input:}}
    \SetKwFunction{FTrain}{Train}
    \SetKwFunction{FExecute}{Plan}
    \SetKwProg{Fn}{Function}{:}{}
    
    % \vspace{1mm}
    
    \KwIn{available dynamics $\hat{\gT}$, 
    % true MDP $\gM = (\gS, \gA, \gT, \pzero, r, \gamma)$,
    % start state distribution $\pzero$, discount factor $\gamma$, 
    expert demonstrations $\taue$, rollout budget $B$, rollout policy $\pi$, horizon $H$,
    % true dynamics $\gT$, 
    test start state $s_0$\\
     \textbf{Note:} For brevity, we omit relevant MDP parameters in the list of arguments.
    }

    \vspace{2mm}
    
    \Fn{\FTrain{$\taue$
    % , $\hat{\gT}$, $\pzero$, $\gamma$ 
    }}{
        Learn a policy $\pitheta$ and a reward function $\rphi$ with any existing IRL algorithm given access to demonstrations $\taue$
        % and MDP parameters  $\gT_{train}$, $\pzero$, and $\gamma$
        , \eg, GAIL
        
        Estimate a value function $\vpsi$ for $\pitheta$
        
        \KwRet $\pitheta$, $\rphi$, $\vpsi$\;
    }
    \vspace{2mm}
    \Fn{\FExecute{$s$, $\pitheta$, $\vpsi$, $\rphi$, $\pi$, $H$, $B$, $\hat{\gT}$
    % , $\pzero$, $\gamma$
    }}{
        Set $s = s_0$
        
        \While{agent is alive} {
            /* Agent planning */
            
            % Propose $B$ actions $\{a^{(1)}, a^{(2)}, ..., a^{(B)}\}$ sampled independently from $\pitheta$\\
            Sample $B$ trajectories $\{\tau^{(1)}, \tau^{(2)}, ..., \tau^{(B)}\}$ of max length $H$ starting from $s$ using dynamics $\hat{\gT}$; sample the first action $a^{(i)}_0 \sim \pitheta$, and sample subsequent actions from $\pi$ as $a^{(i)}_{>0} \sim \pi$, for $i \in \{1, 2, ..., B\}$
            
            % First action sampled from $\pitheta$ and subsequent actions from rollout policy $\pi$. \\
            Estimate trajectory returns $\hat{R}_{\phi, \psi}(\tau^{(i)})$ using $\vpsi$ and $\rphi$
            % , and $\gamma$ 
            (see Eq.~\ref{eq:approx_return})
            
            Pick best action index  $i^\ast = \argmax_i \hat{R}_{\phi, \psi}(\tau^{(i)})$ and execute the best action $a^{(i^\ast)}_0$ 
            
            /* Environment feedback */
            
            Observe true reward $r(s, a^{(i^\ast)}_0)$ and true next state $s \sim \gT(\cdot \vert s, a^{(i^\ast)}_0)$ 
        }
  } 
 \caption{\textit{Imitation with Planning at Test-time} (IMPLANT)}\label{alg:implant}
\end{algorithm}

In the previous section, we showed that imitation learning algorithms based on IRL consider reward inference as an auxiliary task.
% for imitation learning.
Once the agents have been trained, the reward function is discarded and the learned policy is deployed.
% \footnote{In some cases, the reward function is transferred to a new environment and a new policy is learned using the reward function and \textit{additional interactions} with the new environment. See Section~\ref{sec:related} for further discussion.}
Indeed, if the RL step post reward inference (\eg, generator updates in GAIL) were optimal, then the reward function provides no additional information about the expert relative to the imitation policy.
However, this is far from reality, as current RL algorithms can fail to return optimal solutions due to either representational or optimization issues.
For example, there might be a mismatch in the architecture of the policy network and the expert policy, and/or difficulties in optimizing non-convex objective functions.

In fact, the latter challenge gets exacerbated in adversarial learning scenarios due to a non-stationary reward. 
For reference, we consider the policies and discriminator-based rewards learned via GAIL. 
Ideally, we would expect the distribution of inferred discriminator-based rewards for the (state, action) distribution induced by the expert to match the imitation policy (generator) post-training.
We show the empirical distribution of rewards for three MuJoCo environments in Figure~\ref{fig:gail_disc}.
As we can see, there are noticeable differences in the empirical distribution of rewards for the learned agents and the expert, suggesting the challenges in fully exploiting the reward signal inferred through the discriminator simply via policy optimization in the RL step.

\begin{table*}[t]
    \centering
    \caption{Average return of imitation learning algorithms on MuJoCo benchmarks.
    % Number of expert demonstrations in parenthesis.
    }
    \vspace{0.5em}
    \small
    \begin{tabular}{lccc}
    \toprule

     & Hopper & HalfCheetah  & Walker2d  \\
    \midrule
    Expert              & $3570$        & $9892$         & $4585$ \\
    \hline
    BC                  & $127 \pm 85 $ & $-359 \pm 247$ & $153 \pm 162$ \\
    BC-Dropout          & $169 \pm 105$ & $-99 \pm 229$ & $344 \pm 72$ \\
    GAIL                & $3506 \pm 337$& $4059 \pm 728$ & $3847 \pm 635$ \\
    % GAIL-Dropout        & $127 \pm 50$     & $-1 \pm 1$     & $269 \pm 6$ \\
    GAIL-Reward Only      & $319 \pm 123$     & $-284 \pm 77$     & $2 \pm 5$ \\
    IMPLANT (ours)      & $\textbf{3633} \pm 50$ & $\textbf{5240} \pm 924$& $\textbf{4403} \pm 242$ \\
    \bottomrule
    \end{tabular}

    \label{tab:1}
\end{table*}

Building off these observations, we propose \textit{Imitation with Planning at Test-time} (IMPLANT), an imitation learning algorithm that employs the learned reward function for decision-time planning.
The pseudocode for IMPLANT is shown in Algorithm~\ref{alg:implant}.
We can dissect IMPLANT into two sequential phases: a training phase and a planning phase. 

\paragraph{Training phase:} We can invoke any imitation learning algorithm, \eg, GAIL, that optimizes for a stochastic imitation policy $\pitheta$ to maximize some inferred reward function $r_\phi$.
Given the challenges due to non-identifiability of the true reward function in imitation learning~\citep{ng2000algorithms}, the inferred reward function is typically only a proxy signal for learning e.g., discriminator outputs in adversarial methods~\citep{ho2016generative,fu2017learning}, unsupervised perceptual rewards for self-supervised imitation learning~\citep{sermanet2016unsupervised}, etc.

We also train a parameterized value function $V_\psi$ at this stage. 
Value function estimation is often a subroutine for many RL algorithms including those used to update the policy within the IRL setup, such as actor-critic methods~\citep{konda2000actor}.
For such algorithms, learning a value function does not incur any additional computation.

% such as actor-critic methods,  updating the policy within the IRL 

\paragraph{Planning phase:} At decision-time, we use the imitation policy along with the learned value and reward functions for closed-loop planning. 
We build our planner based on model-predictive control (MPC)~\citep{camacho2013model}.
At any given state $s_t$ and time $t\geq 0$, we are interested in choosing action sequences for trajectories which maximizes the following objective:
\begin{align}\label{eq:mpc}
    % a_t, a_{t+1}, \cdots, a_{t+H-1} = \argmax_{a_t, a_{t+1}, \cdots, a_{t+H-1}} R(\tau) = \sum_{t'=t}^{t+H-1} \gamma^{t'} r(s_{t'}, a_{t'})
    a_t, a_{t+1}, \cdots, = \argmax_{a_t, a_{t+1}, \cdots} R(\tau) = \sum_{t'=t}^{\infty} \gamma^{t'-t} r(s_{t'}, a_{t'})
\end{align}
where $s_0 \sim \pzero$ and $s_{t+1} \sim \gT(\cdot \vert s_t, a_t)$ for all $t\geq 0$. 

This objective has also been applied for model-based RL with a learned dynamics model and black-box access to the rewards function~\citep{nagabandi2018neural,chua2018deep}.
Unlike the RL setting, however, we do not know the reward function for imitation learning.
We will assume access to some approximation of the true dynamics such as a simulator~\citep{ho2016generative} or a model estimated from expert demonstrations and/or training interactions~\citep{baram2016model}.
As we shall demonstrate in our experiments, IMPLANT is robust even when the true dynamics $\gT$ perturb from the dynamics available for planning $\hat{\gT}$. 
% The true dynamics model may be available for planning (\ie, $\hat{\gT}=\gT$) as in \citet{ho2016generative} or can be estimated from expert demonstrations or online interactions~\citep{baram2016model}.
Given $\hat{\gT}$, we can do rollouts as before in regular model-based RL but need to rely on learned estimates for the reward function.
In particular, we use the learned reward function $\rphi$ up to a fixed horizon $H$ and a terminal value function $\vpsi$ thereafter to estimate the trajectory return as:
\begin{align}\label{eq:approx_return}
    R(\tau) \approx \sum_{t'=t}^{t+H-1} \gamma^{t'-t} \rphi(s_{t'}, a_{t'}) + \gamma^{H} \vpsi(s_{H}):=  \hat{R}_{\phi, \psi}(\tau).
\end{align}

Substituting Eq.~\ref{eq:approx_return} in Eq.~\ref{eq:mpc}, we obtain a surrogate objective for optimization. 
To optimize this surrogate, we propose a variant of the random shooting optimizer~\citep{richards2005robust} that works as follows.
At the current state $s_t$, we first  sample a set of $B$ candidate actions independently from the imitation policy.
For each candidate action, we estimate a score based on their expected returns by performing rollout(s) of fixed-length $H$.
The rollout policy $\pi$ from which we sample all subsequent actions could be random (potentially high variance) or the imitation policy $\pitheta$ (potentially high bias) or a mixture. 
In our experiments, we obtained consistently better performance with using $\pitheta$ as the rollout policy $\pi$ and taking the mean of $\pi$ instead of sampling.
For each trajectory, we estimate its return via Eq.~\ref{eq:approx_return} and finally, pick the action with the largest return.

Consistent with the closed-loop nature of MPC, we repeat the above procedure at the next state $s_t$. 
Doing so helps correct for errors in estimation and optimization in the previous time step, albeit at the expense of additional computation.
The algorithm has two critical parameters that induce similar computational trade-offs. First, we need to specify a budget $B$ for the total number of rollouts.
The higher the budget, larger is our search space for the best action.
Second, we need to specify a planning horizon $H$.
For larger lengths, we need extra computation to interact with the dynamics of the environment and rely more on the learned reward function than the value function for estimating returns in Eq.~\ref{eq:approx_return}.
However, since the rollouts are independent, we can mitigate additional computational costs by parallelizing them. While this parallelization is indeed bottlenecked by the last finished rollout, in all of our experiments, we perform rollouts of fixed length and the optimal horizon is relatively small $(10\sim50)$. Thus, the gains due to parallelization are significant.

\section{Experiments}
% old values
% \begin{table*}
%     \centering
%     \caption{Average return of imitation learning algorithms in causal confusion setting.}
%     \vspace{0.5em}
%     \small
%     \begin{tabular}{lccc}
%     \toprule

%      & Hopper & HalfCheetah & Walker2d \\
%     \midrule
%     Expert              & $3570$        & $891$         & $3593$ \\
%     \hline
%     BC                  & $209 \pm 121$ & $331 \pm 141$ & $119 \pm 206$ \\
%     BC-Dropout         & $162 \pm 108$ & $548 \pm 175$ & $700 \pm 433$ \\
%     GAIL                & $579 \pm 484 $ & $699 \pm 200$ & $613 \pm 465$ \\
%     % GAIL-Dropout        & $63 \pm 6$     & $-6 \pm 4$     & $272 \pm 16$ \\
%     GAIL-Reward Only      & $515 \pm 302$     & $-82 \pm 79$     & $57 \pm 146$ \\
%     IMPLANT (ours)      & $\textbf{1717} \pm 1262$ & $\textbf{827} \pm 375$& $\textbf{807} \pm 395$ \\
%     \bottomrule
%     \end{tabular}

%     \label{tab:2}
% \end{table*}

Our experiments aim to evaluate IMPLANT in two kinds of settings.
First, we evaluate its performance in the default ``no-transfer" setting, where the agent is trained and tested in the same environment.
Second, we emphasize the robustness of IMPLANT by evaluating its zero-shot generalization performance in environments where the test dynamics are a perturbed version of the training dynamics. 
We consider three such perturbations: causal confusion~\citep{de2019causal}, motor noise~\citep{christiano2016transfer}, and transition noise.
We will describe each of these setups subsequently alongside the results. For all transfer settings, we only assume access to the training simulator and use it as $\hat{\gT}$ for planning. At test-time, no additional interactions is allowed, nor do we have access to the test dynamics $\gT_{test}$.

\begin{table*}[ht]
    \centering
        \caption{Average return of imitation learning algorithms in causal confusion: action nuisance.}
    \label{tab:2}
    % \small
    \begin{tabular}{lccc}
    \toprule

     & Hopper & HalfCheetah & Walker2d \\
    \midrule
    Expert              & $3570$        & $9892$         & $4585$ \\
    \hline
    BC                  & $209 \pm 121$ & $-297 \pm 308$ & $160 \pm 135$ \\
    BC-Dropout         & $162 \pm 108$ & $-159 \pm 300$ & $230 \pm 120$ \\
    GAIL                & $579 \pm 484 $ & $1367 \pm 660$ & $2052 \pm 745$ \\
    % GAIL-Dropout        & $63 \pm 6$     & $-6 \pm 4$     & $272 \pm 16$ \\
    GAIL-Reward Only      & $515 \pm 302$     & $-90 \pm 90$     & $114 \pm 191$ \\
    IMPLANT (ours)      & $\textbf{1717} \pm 1262$ & $\textbf{3890} \pm 1080$& $\textbf{3891} \pm 585$ \\
    \bottomrule
    \end{tabular}
\end{table*}
\begin{table*}[htbp]
    \centering
        \caption{Average return of imitation learning algorithms in causal confusion: state nuisance.}
    \label{tab:2b}
    \centering

    % \small
    \begin{tabular}{lccc}
    \toprule

     & Hopper & HalfCheetah & Walker2d \\
    \midrule
    Expert              & $3570$        & $9892$         & $4585$ \\
    \hline
    BC                  & $203 \pm 115$ & $-409 \pm 233$ & $91 \pm 175$ \\
    BC-Dropout         & $213 \pm 135$ & $-207 \pm 229$ & $422 \pm 85$ \\
    GAIL                & $1733 \pm 1402 $ & $2492 \pm 748$ & $2628 \pm 1237$ \\
    % GAIL-Dropout        & $63 \pm 6$     & $-6 \pm 4$     & $272 \pm 16$ \\
    GAIL-Reward Only      & $362 \pm 41$     & $-38 \pm 131$     & $125 \pm 197$ \\
    IMPLANT (ours)      & $\textbf{3642} \pm 85$ & $\textbf{5465} \pm 889$& $\textbf{4527} \pm 308$ \\
    \bottomrule
    \end{tabular}
\end{table*}

\paragraph{Setup.} We evaluate our approach on MuJoCo enviroments in OpenAI Gym~\citep{brockman2016openai}: Hopper, HalfCheetah, and Walker2d. We obtain the expert data by training a SAC agent~\citep{haarnoja2018soft}.
% and record its rollouts.
% The expert data used for benchmarking imitation learning on these environments is publicly available\footnote{\url{https://github.com/openai/baselines}}.
We replicate the experimental setup of \citet{ho2016generative} by fixing a limited number of expert trajectories used for training, as well as sub-sampling expert trajectories every 20 time steps.
All results are averaged over 5 runs of each algorithm with different seeds.
 We provide further details in Appendix~\ref{ap:1}.
% constructing 3 challenging setups where the test dynamics

% answer the following questions: 
% \begin{enumerate}
%   \item Does IMPLANT outperform GAIL algorithm?
%   \item Is IMPLANT better at zero-shot generalization when there are changes in test-time dynamics due to ``causal confusion''?
%   \item Is IMPLANT better at zero-shot generalization when there are perturbations in test-time dynamics?
%   \item How do various design choices affect IMPLANT?
% \end{enumerate}

% \paragraph{Setup.} We now evaluate our algorithm against baselines on 3 of the MuJoCo \citep{todorov2012mujoco} environments: Hopper, HalfCheetah, and Walker2d. 
% Our expert data comes from OpenAI's Baselines github repository \citep{openai2017baselines}. We replicate the experimental setup of \citet{ho2016generative} by fixing a limited number of expert trajectories used for training, as well as sub-sampling expert trajectories every 20 time steps starting with a random offset. For GAIL, we use trust region policy optimization \citep{schulman2015trust} as our policy optimization algorithm. More of our setup information and the hyperparameter choices are detailed in Appendix \ref{ap:2}. All results are averaged over 5 seeds.

\paragraph{Baselines.} As we observed in Algorithm~\ref{alg:implant}, IMPLANT can employ any IRL algorithm under the hood. 
For our experiments, we consider GAIL~\citep{ho2016generative} as the IRL algorithm of choice both as input for IMPLANT and consequently, as the closest baseline of interest.
GAIL is amongst the current state-of-the-art methods for imitation learning; see Section~\ref{sec:irl} for a detailed description.
For every environment, we report results for IMPLANT using a single set of hyperparameters for the rollout budget and planning horizon. We provide further details in Appendix~\ref{ap:1}.

Additionally, we consider a Behavioral Cloning (BC) baseline; see Section~\ref{sec:bc} for a detailed description.
We also tested variants of GAIL and BC with dropout~\citep{srivastava2014dropout} to demonstrate the limited utility of standard regularization techniques in countering the challenges due to low data and test noise.
In fact, GAIL with dropout completely failed to learn in the adversarial setting on all environments; for brevity, we exclude it from presentation. 

Last, we include a ``GAIL-Reward Only" ablation baseline where we discard the imitation policy (generator) of GAIL during execution and instead, only use the inferred reward model (discriminator) in conjunction with a random policy for decision-time planning.  
This directly contrasts with the GAIL baseline, which by default only uses the generator.
On the other hand, IMPLANT uses both the generator and discriminator for imitation via decision-time planning.

% The IL agents that we compare our model against are Behavioral Cloning, Behavioral Cloning with Dropout , GAIL, and GAIL with Dropout. All of the baseline agents and our model are trained with the following setup.

\subsection{Imitation with Limited Expert Trajectories}\label{sec:exp_default}

With a very low number of expert trajectories, 
% it has been shown by 
\citet{ho2016generative} showed that GAIL achieves excellent performance in almost all these environments.
% We strength test its performance by shrinking the expert trajectories to almost half.
% We evaluate the performance of IMPLANT using the lowest number of expert trajectories tested in prior work.
In the first set of experiments, we test IMPLANT under the same constraints to estimate its raw performance (without any test-time perturbations).
The results are shown in Table~\ref{tab:1}.
We find that IMPLANT consistently outperforms existing algorithms on all environments. It also achieves near-optimal performance, with the exception of HalfCheetah,  where even GAIL performs noticeably worse than the expert.
As expected, BC and BC-Dropout perform poorly in this setting. 
GAIL-Reward Only exhibits the poorest performance suggesting the benefits of explicitly learning a parametric policy.

% and GAIL can still reach near optimal performance , with 

% Table \ref{tab:1} shows that our algorithm outperforms all of the baseline imitation learning algorithms. We see that BC and BC-Dropout suffers greatly from limited expert data, whereas IMPLANT and GAIL can still reach near optimal performance with such constraints. Unfortunately, GAIL with dropout cannot learn a meaningful policy in our setup.

\subsection{Zero-Shot Transfer: Causal Confusion} \label{exp:causal}

 \citet{de2019causal} observed that imitation learning approaches are susceptible to \textit{causal confusion}, \ie, their performance deteriorates significantly in the presence of nuisance confounders in the state representation.
 To demonstrate this phenomena empirically, \citet{de2019causal} further propose a challenging setup in which the nuisance is created by appending the agent's observation with its action from the previous time step.
 A standard imitation agent trained in this setup will learn to \textit{copy} the previous action (since successive actions are highly correlated in expert demonstrations), falling prey to causal confusion.
 At test-time, the agent's performance drops drastically if the appended action is replaced by random noise (\ie, the confounding is removed).
 We refer the reader to \citet{de2019causal} for further details and analysis.
 %  Following \citet{de2019causal}, 
We benchmark IMPLANT and other imitation learning algorithms under two causal confusion setups: 
\begin{enumerate}
    \item  \textbf{Action nuisance}. This follows \citet{de2019causal}, where the action from the previous step is appended to the observation to create nuisance. \\
\item \textbf{State nuisance}. Alternatively, we can corrupt the observation directly by appending a nuisance variable $s$ to the observation that correlates with high reward in the training environment but is absent at test time.
 For example, in locomotion environments such as Hopper, when the agent crosses a certain velocity threshold ($v \geq v_{th}$), we set $s=1$. At test time, the nuisance is disabled completely i.e., $s=0$.
 \end{enumerate}

%  Motivated by the car example in \citet{de2019causal}, we add a signal $s$ to the observation, which turns on ($s=1$) when the agent reaches certain velocity threshold ($v \geq v_{th}$). In the test environment, we disable the signal completely ($s=0$).
 
\begin{figure*}[htbp]
    \centering
    \begin{subfigure}[b]{0.9\textwidth}
        \centering
        \begin{subfigure}[b]{0.48\textwidth}
        \centering
        \begin{subfigure}[b]{0.18\textwidth}
        \centering
        \includegraphics[width=\textwidth]{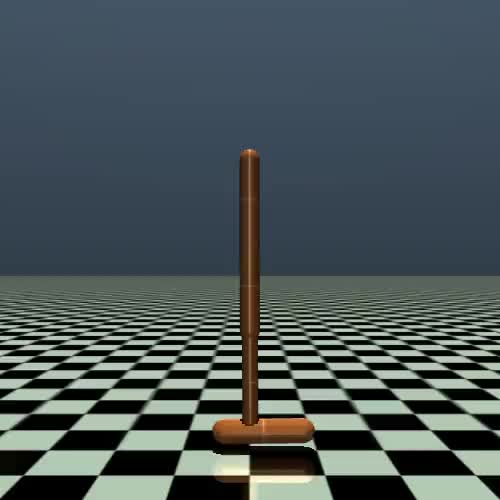}
        \end{subfigure}
        \begin{subfigure}[b]{0.18\textwidth}
        \centering
        \includegraphics[width=\textwidth]{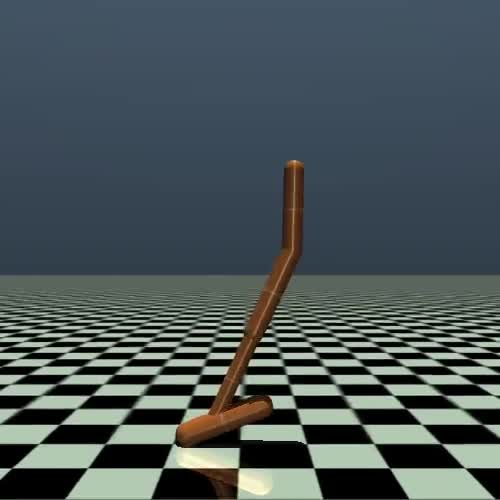}
        \end{subfigure}
        \begin{subfigure}[b]{0.18\textwidth}
        \centering
        \includegraphics[width=\textwidth]{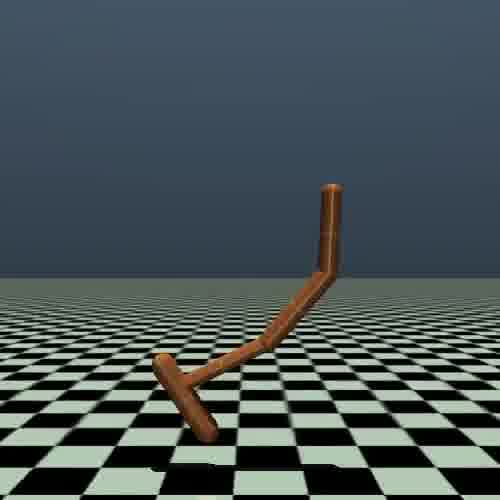}
        \end{subfigure}
        \begin{subfigure}[b]{0.18\textwidth}
        \centering
        \includegraphics[width=\textwidth]{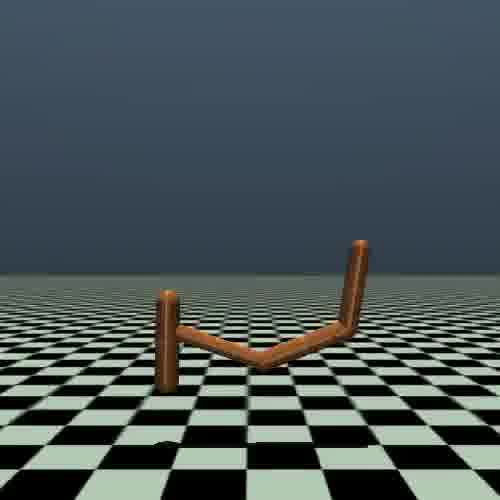}
        \end{subfigure}
        \begin{subfigure}[b]{0.18\textwidth}
        \centering
        \includegraphics[width=\textwidth]{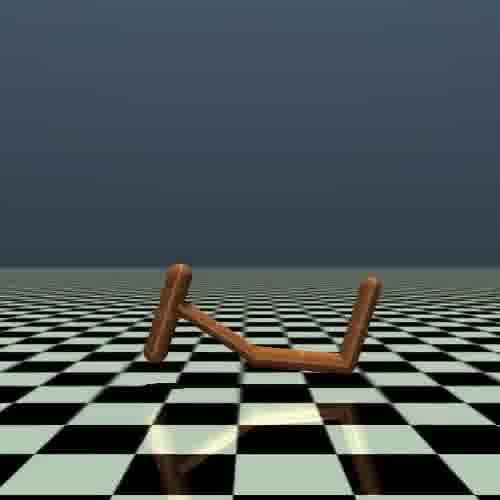}
        \end{subfigure}
        \caption{Hopper: GAIL}
        \label{fig3:gail-hopper}
        \end{subfigure}
        \hspace{2.5mm} 
        \centering
        \begin{subfigure}[b]{0.48\textwidth}
            \centering
            \begin{subfigure}[b]{0.18\textwidth}
            \centering
            \includegraphics[width=\textwidth]{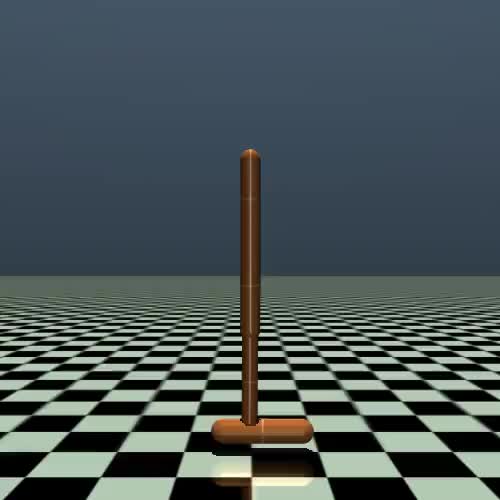}
            \end{subfigure}
            \begin{subfigure}[b]{0.18\textwidth}
            \centering
            \includegraphics[width=\textwidth]{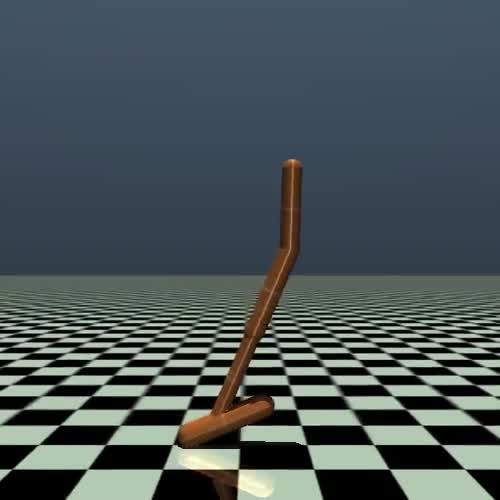}
            \end{subfigure}
            \begin{subfigure}[b]{0.18\textwidth}
            \centering
            \includegraphics[width=\textwidth]{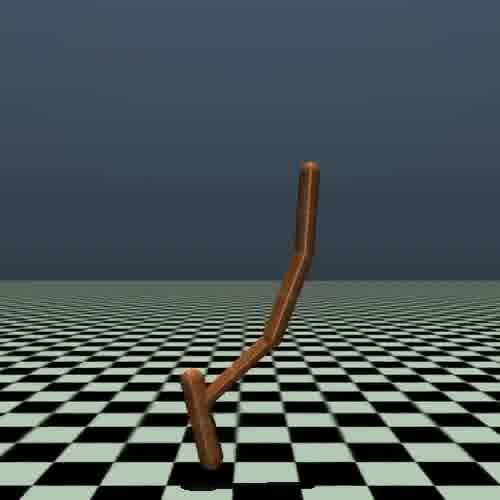}
            \end{subfigure}
            \begin{subfigure}[b]{0.18\textwidth}
            \centering
            \includegraphics[width=\textwidth]{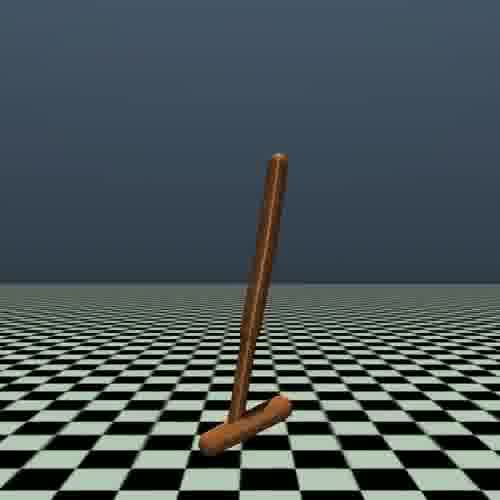}
            \end{subfigure}
            \begin{subfigure}[b]{0.18\textwidth}
            \centering
            \includegraphics[width=\textwidth]{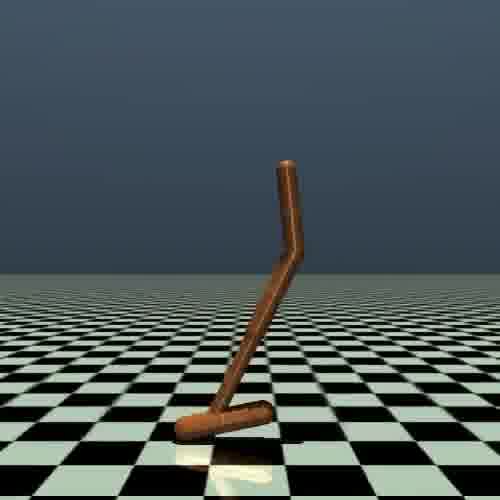}
            \end{subfigure}
            \caption{Hopper: IMPLANT}
            \label{fig3:implant-hopper}
        \end{subfigure}
    \end{subfigure}
    
    \begin{subfigure}[b]{0.9\textwidth}
        \centering
        \begin{subfigure}[b]{0.48\textwidth}
        \centering
        \begin{subfigure}[b]{0.18\textwidth}
        \centering
        \includegraphics[width=\textwidth]{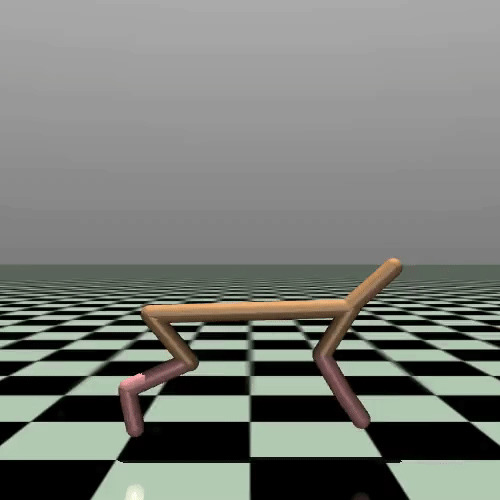}
        \end{subfigure}
        \begin{subfigure}[b]{0.18\textwidth}
        \centering
        \includegraphics[width=\textwidth]{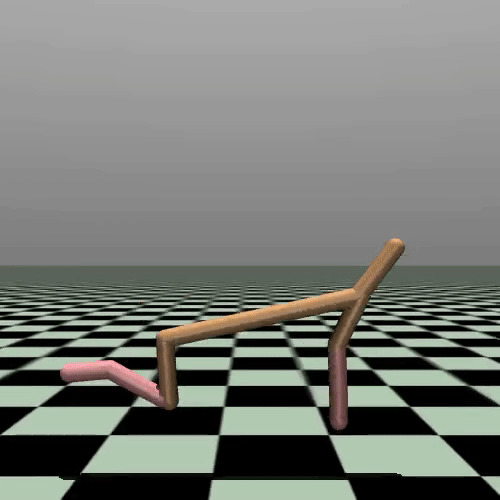}
        \end{subfigure}
        \begin{subfigure}[b]{0.18\textwidth}
        \centering
        \includegraphics[width=\textwidth]{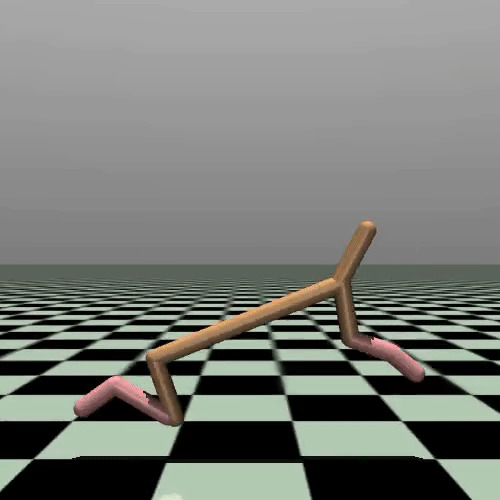}
        \end{subfigure}
        \begin{subfigure}[b]{0.18\textwidth}
        \centering
        \includegraphics[width=\textwidth]{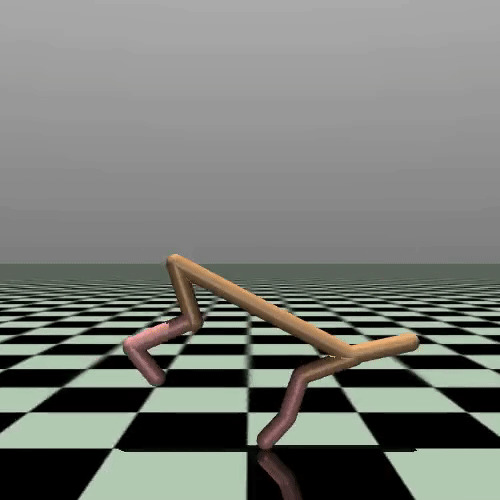}
        \end{subfigure}
        \begin{subfigure}[b]{0.18\textwidth}
        \centering
        \includegraphics[width=\textwidth]{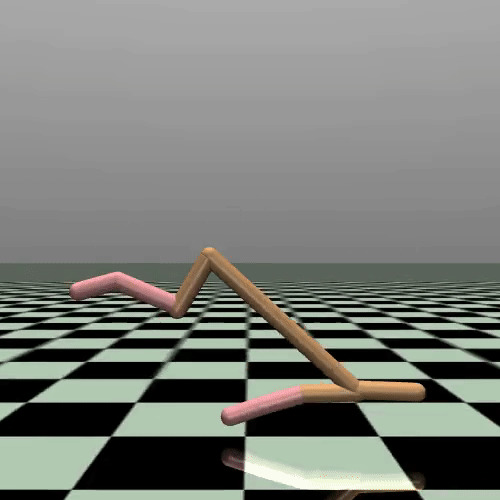}
        \end{subfigure}
        \caption{HalfCheetah: GAIL}
        \label{fig3:gail-cheetah}
        \end{subfigure}
        \hspace{2.5mm}
        \centering
        \begin{subfigure}[b]{0.48\textwidth}
            \centering
            \begin{subfigure}[b]{0.18\textwidth}
            \centering
            \includegraphics[width=\textwidth]{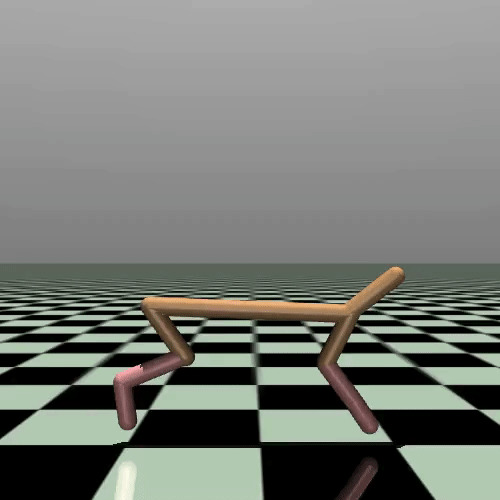}
            \end{subfigure}
            \begin{subfigure}[b]{0.18\textwidth}
            \centering
            \includegraphics[width=\textwidth]{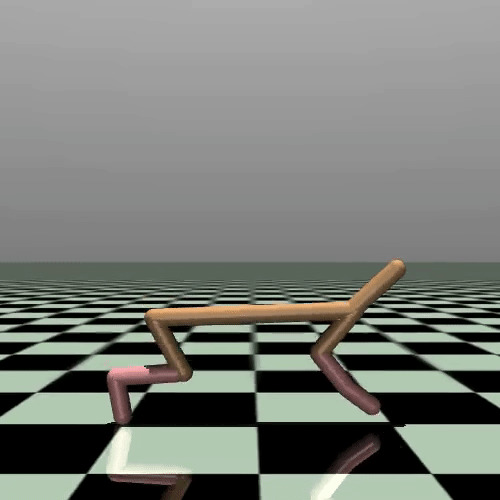}
            \end{subfigure}
            \begin{subfigure}[b]{0.18\textwidth}
            \centering
            \includegraphics[width=\textwidth]{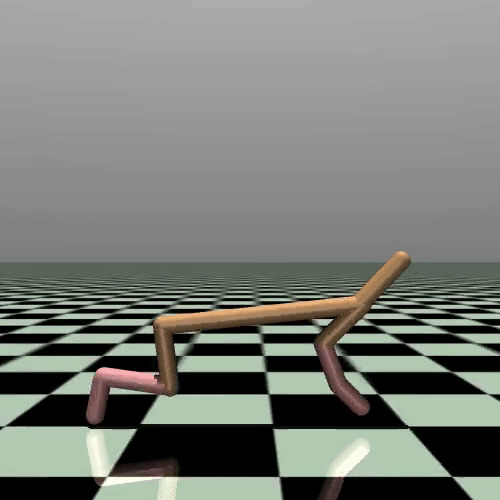}
            \end{subfigure}
            \begin{subfigure}[b]{0.18\textwidth}
            \centering
            \includegraphics[width=\textwidth]{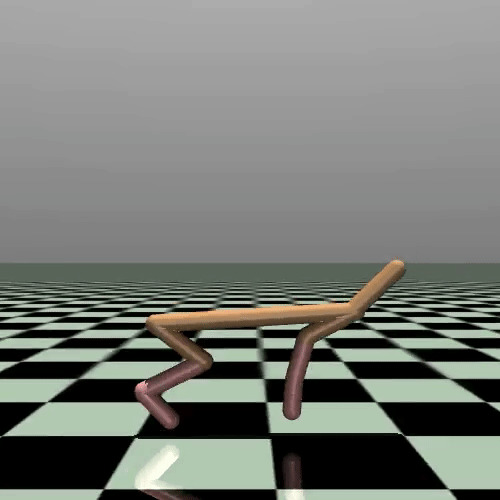}
            \end{subfigure}
            \begin{subfigure}[b]{0.18\textwidth}
            \centering
            \includegraphics[width=\textwidth]{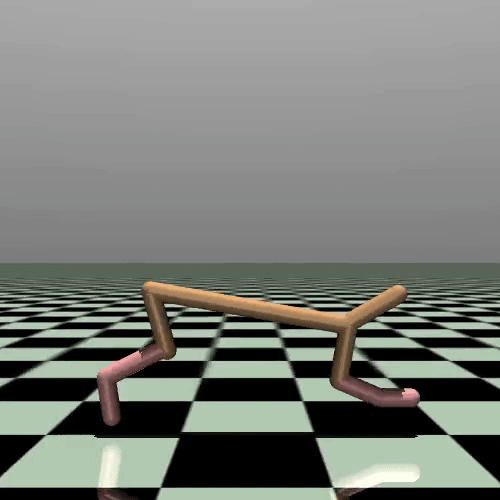}
            \end{subfigure}
            \caption{HalfCheetah: IMPLANT}
            \label{fig3:implant-cheetah}
        \end{subfigure}
    \end{subfigure}
    
    \begin{subfigure}[b]{0.9\textwidth}
        \centering
        \begin{subfigure}[b]{0.48\textwidth}
        \centering
        \begin{subfigure}[b]{0.18\textwidth}
        \centering
        \includegraphics[width=\textwidth]{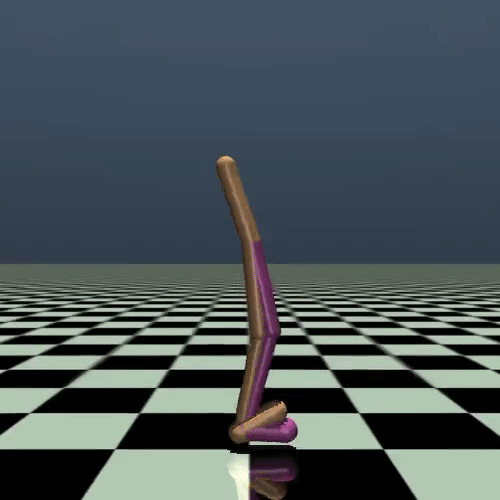}
        \end{subfigure}
        \begin{subfigure}[b]{0.18\textwidth}
        \centering
        \includegraphics[width=\textwidth]{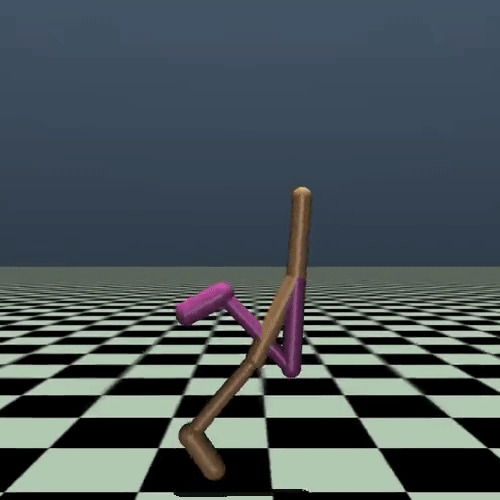}
        \end{subfigure}
        \begin{subfigure}[b]{0.18\textwidth}
        \centering
        \includegraphics[width=\textwidth]{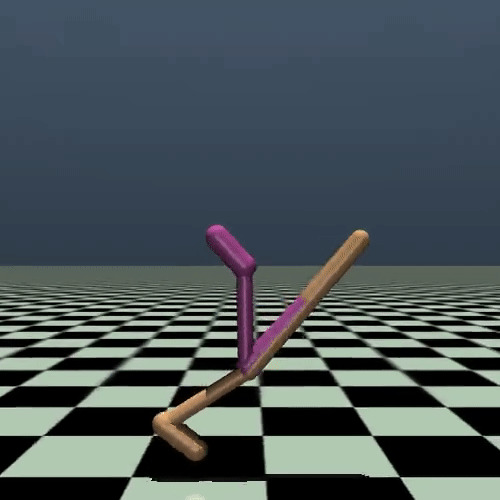}
        \end{subfigure}
        \begin{subfigure}[b]{0.18\textwidth}
        \centering
        \includegraphics[width=\textwidth]{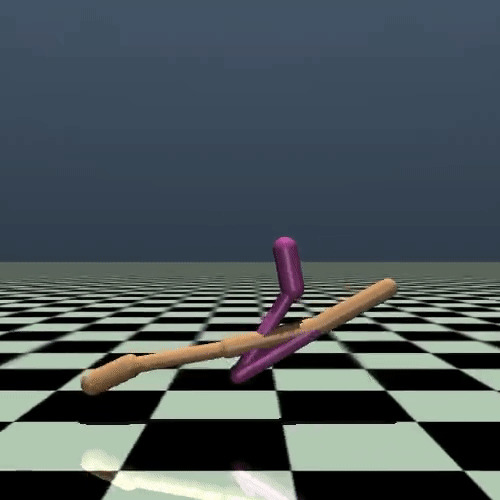}
        \end{subfigure}
        \begin{subfigure}[b]{0.18\textwidth}
        \centering
        \includegraphics[width=\textwidth]{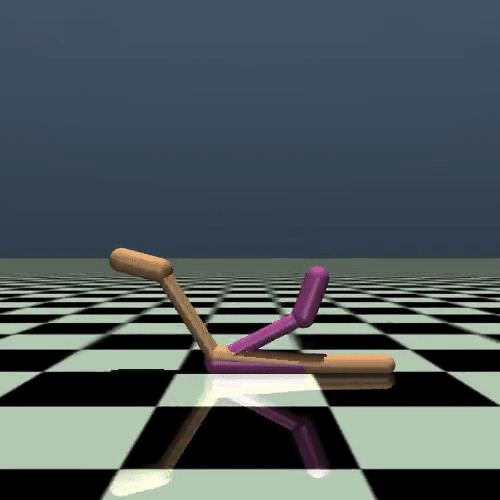}
        \end{subfigure}
        \caption{Walker2d: GAIL}
        \label{fig3:gail-walker}
        \end{subfigure}
        \hspace{2.5mm}
        \centering
        \begin{subfigure}[b]{0.48\textwidth}
            \centering
            \begin{subfigure}[b]{0.18\textwidth}
            \centering
            \includegraphics[width=\textwidth]{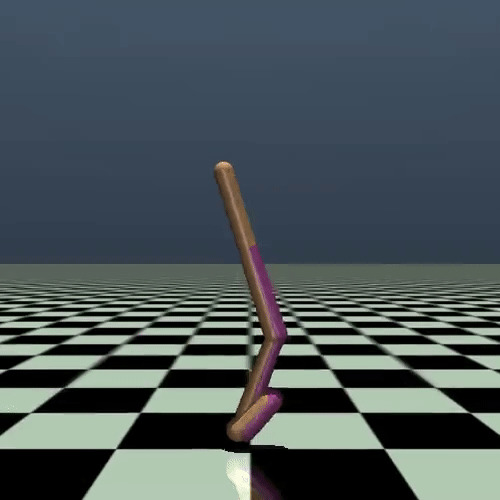}
            \end{subfigure}
            \begin{subfigure}[b]{0.18\textwidth}
            \centering
            \includegraphics[width=\textwidth]{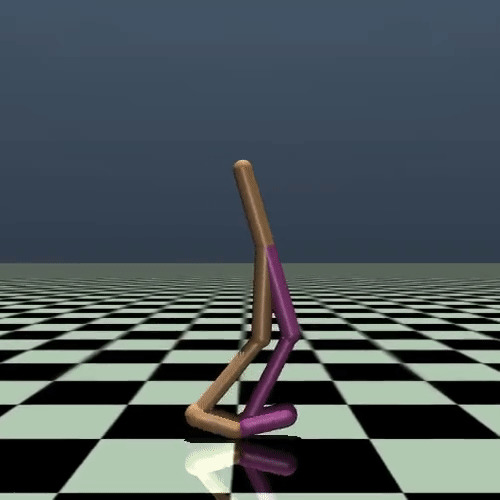}
            \end{subfigure}
            \begin{subfigure}[b]{0.18\textwidth}
            \centering
            \includegraphics[width=\textwidth]{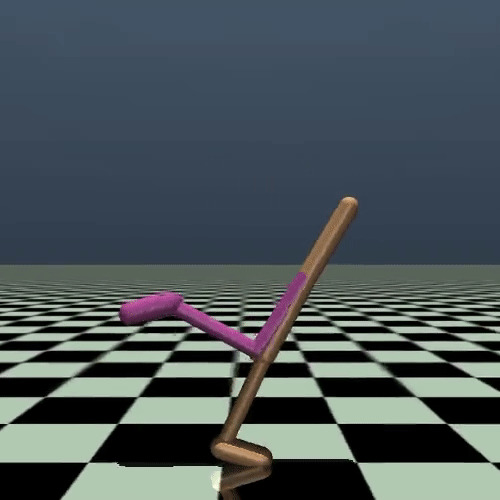}
            \end{subfigure}
            \begin{subfigure}[b]{0.18\textwidth}
            \centering
            \includegraphics[width=\textwidth]{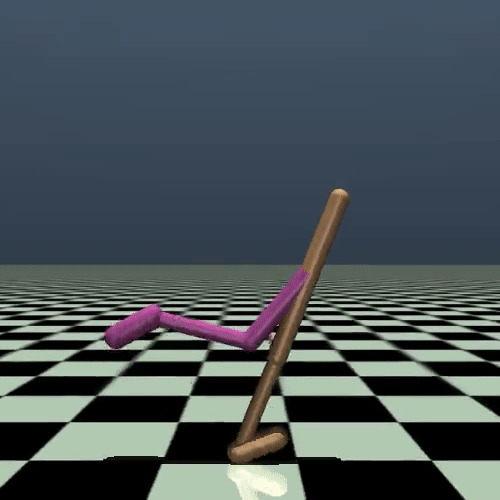}
            \end{subfigure}
            \begin{subfigure}[b]{0.18\textwidth}
            \centering
            \includegraphics[width=\textwidth]{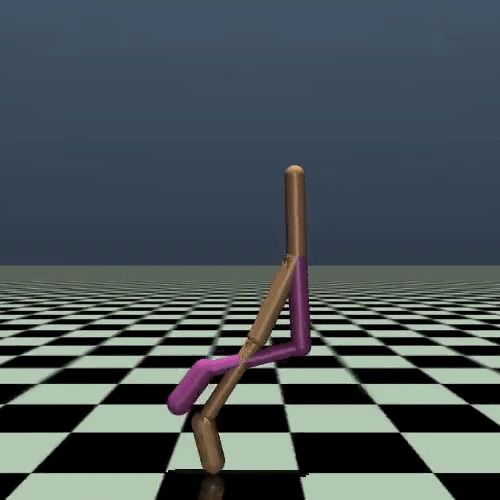}
            \end{subfigure}
            \caption{Walker2d: IMPLANT}
            \label{fig3:implant-walker}
        \end{subfigure}
    \end{subfigure}

    % \begin{subfigure}{0.7\textwidth}
    % \centering
    % \includegraphics[width=\textwidth]{figs/comparison3.png}
    % \caption{Return over time of the above agents}
    % \end{subfigure}
    
    \caption{Trajectory visualization for causal confusion: state nuisance at test-time. While all agents start from the same state (see first frame), only IMPLANT can effectively move forward. All agents are trained in the confounded setting and tested in the non-confounded setting.}
    \label{fig:2}
\end{figure*}

The results are shown in Table~\ref{tab:2} and Table~\ref{tab:2b}.
While all baselines, including GAIL, fail due to the confounding nuisance, IMPLANT is significantly more robust in all environments. Impressively, IMPLANT is able to completely overcome the nuisance and recover the non-confounded policy performance (Table~\ref{tab:1}) in all 3 environments with state nuisance.
We can visualize the agent performance qualitatively in Figure~\ref{fig:2}.
Note that we provided the IMPLANT agent access to only the confounded dynamics for decision-time planning. The algorithm is zero-shot, unlike the proposed solutions of \citet{fu2017learning} and \citet{de2019causal} which require further interactions with the non-confounded test environment for recovery.
 
%  Finally, the authors propose to use target interventions by querying the expert at additional 
%  the agent's observation with an action executed from the previous time step 
% To test whether IMPLANT is robust to nuisance variables in the state representation, we create two settings resembling the setup of \citet{de2019causal}. The states that an agent observes will be augmented with the agent's previous actions or random noise. We will refer to the first setting as ``confounded'' and the second ``original''. One distinction we have from \citet{de2019causal} is we are interested in zero-shot generalization, whereas they allow their agent to perform targeted interventions in test environments, thus not applicable in our setup. We train our agents in the confounded setting and test them in the original setting to see how well they can disentangle the nuisance variables at test-time. For comparison, we also add in a GAIL agent both trained and tested in the confounded setting. Table \ref{tab:2} shows that our method is robust to nuisance variables in the states and it outperforms all the other baselines by a wide margin.

\begin{figure*}[t]
    \centering
    \begin{subfigure}[b]{0.32\textwidth}
    \centering
    \includegraphics[width=\textwidth]{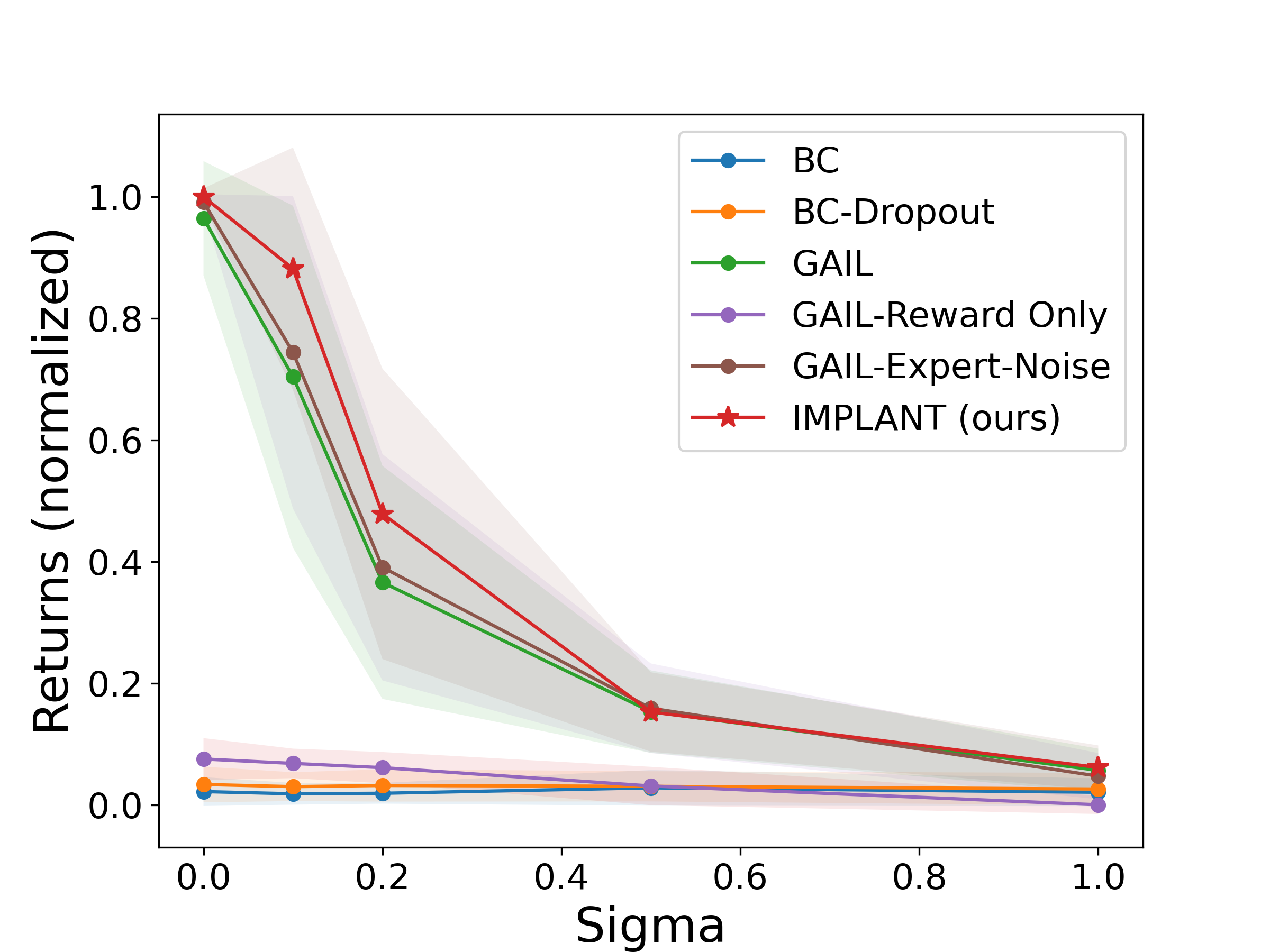}
    \caption{Hopper: Motor Noise}
    \end{subfigure}
    \begin{subfigure}[b]{0.32\textwidth}
    \centering
    \includegraphics[width=\textwidth]{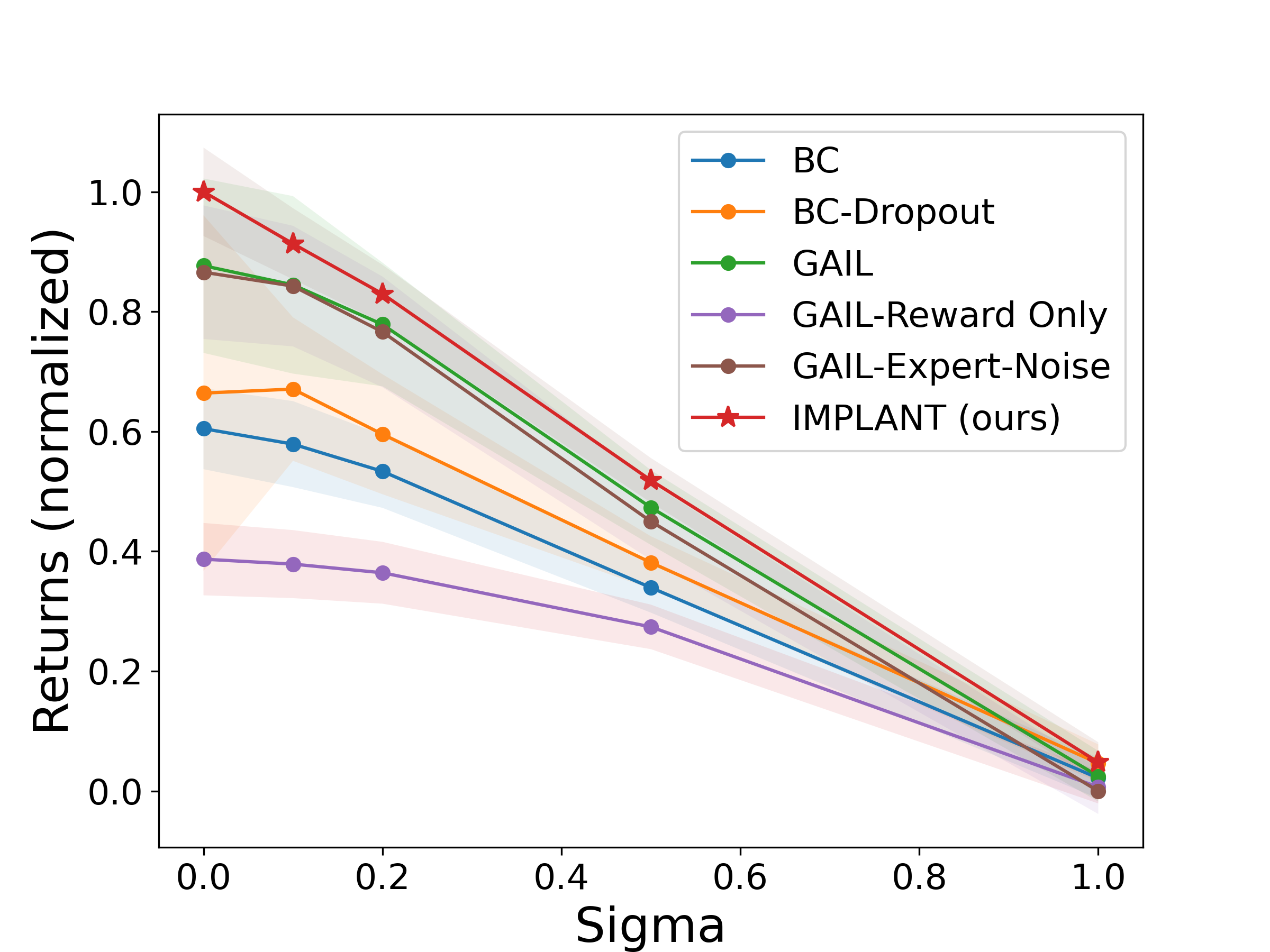}
    \caption{HalfCheetah: Motor Noise}
    \end{subfigure}
    \begin{subfigure}[b]{0.32\textwidth}
    \centering
    \includegraphics[width=\textwidth]{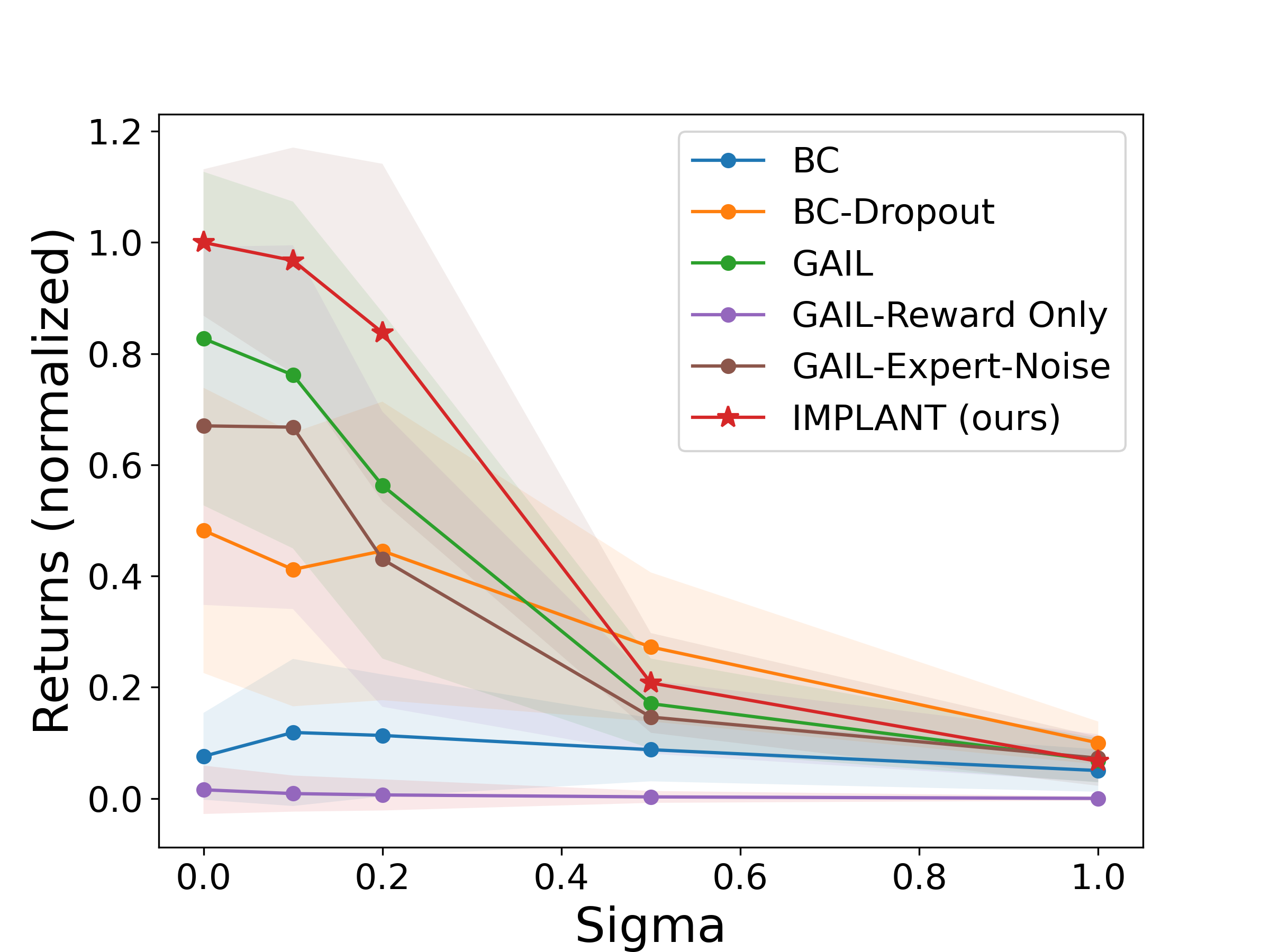}
    \caption{Walker2d: Motor Noise}
    \end{subfigure}
    \begin{subfigure}[b]{0.32\textwidth}
    \centering
    \includegraphics[width=\textwidth]{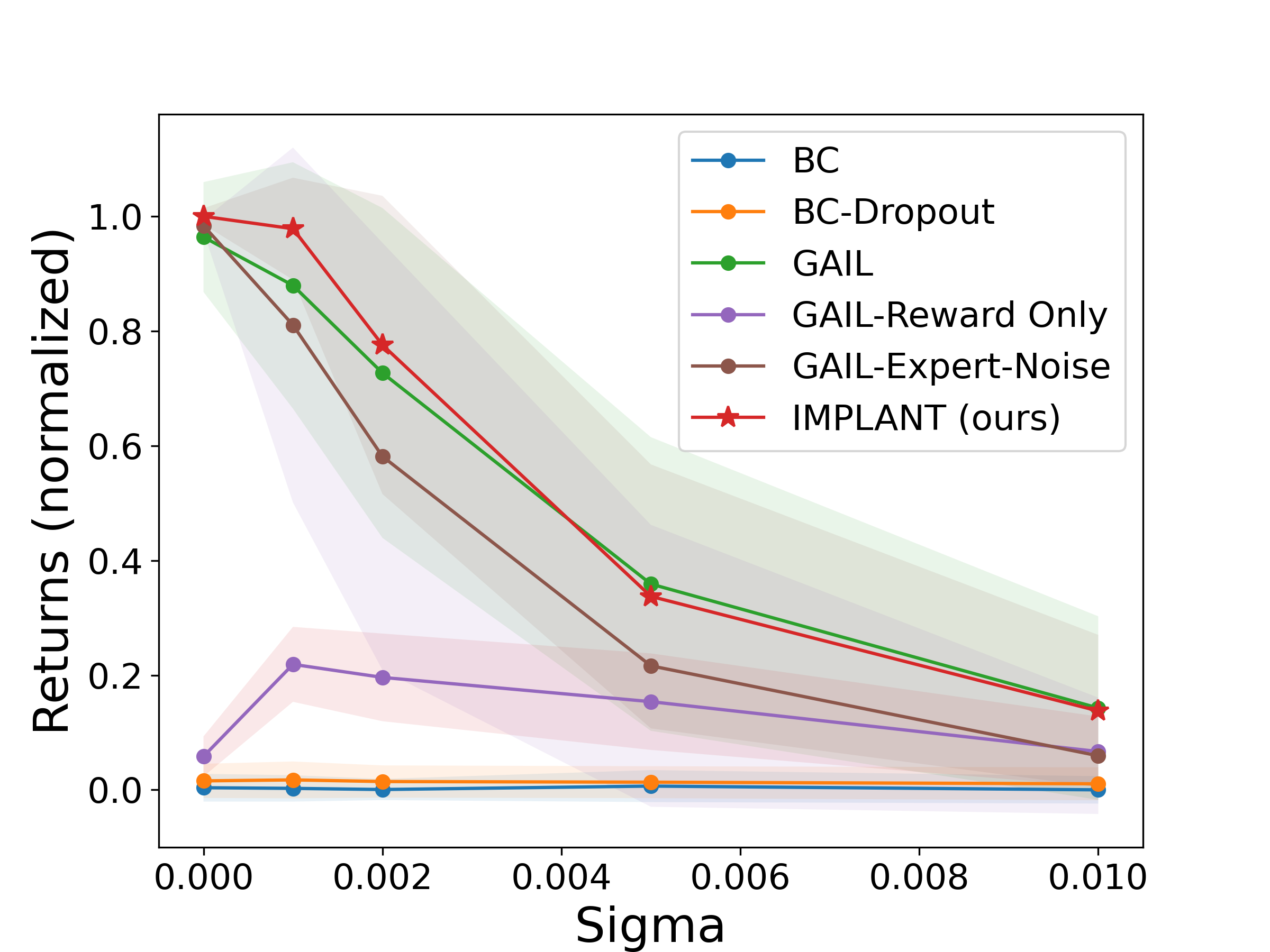}
    \caption{Hopper: Transition Noise}
    \end{subfigure}
    \begin{subfigure}[b]{0.32\textwidth}
    \centering
    \includegraphics[width=\textwidth]{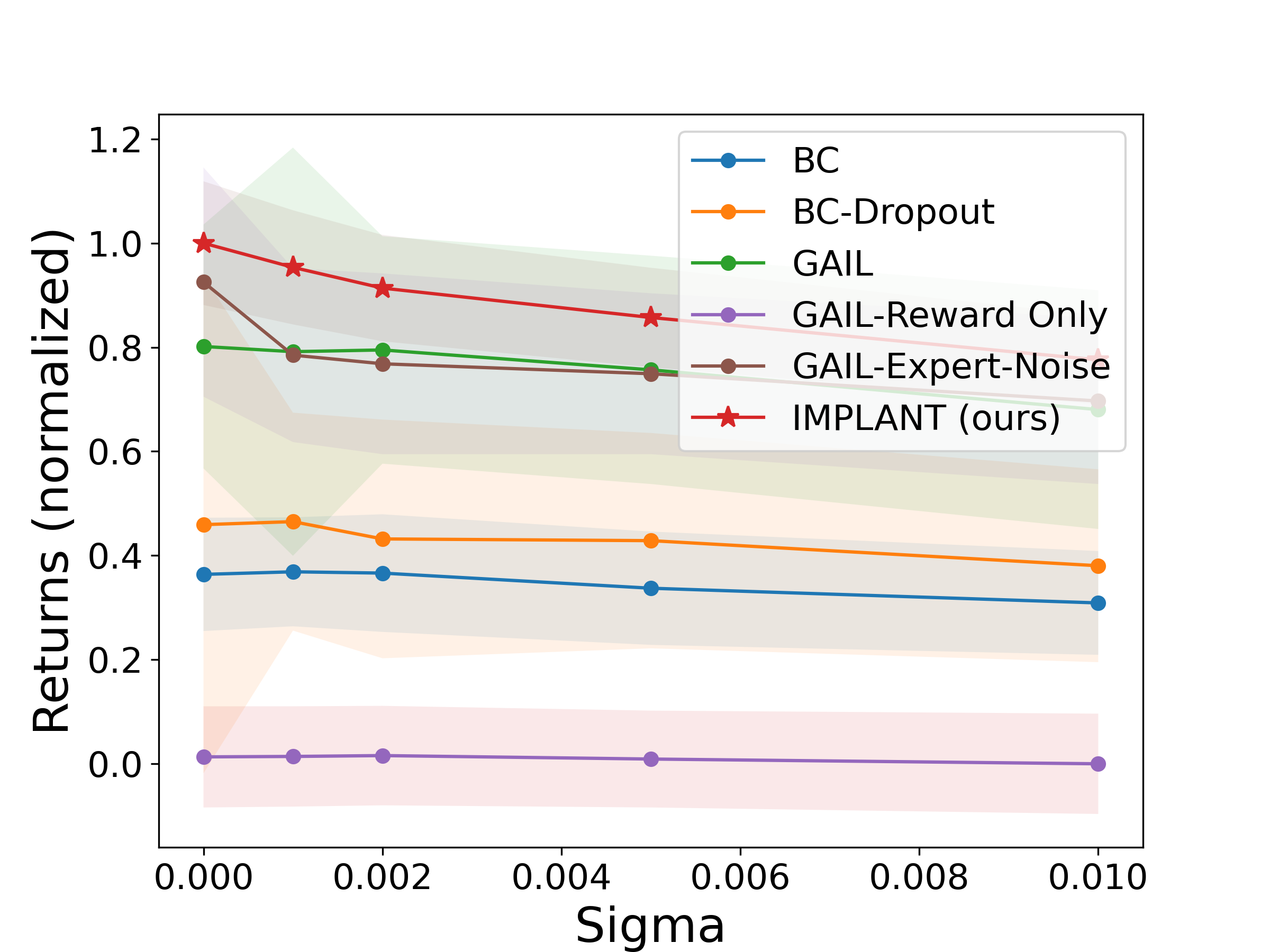}
    \caption{HalfCheetah: Transition Noise}
    \end{subfigure}
    \begin{subfigure}[b]{0.32\textwidth}
    \centering
    \includegraphics[width=\textwidth]{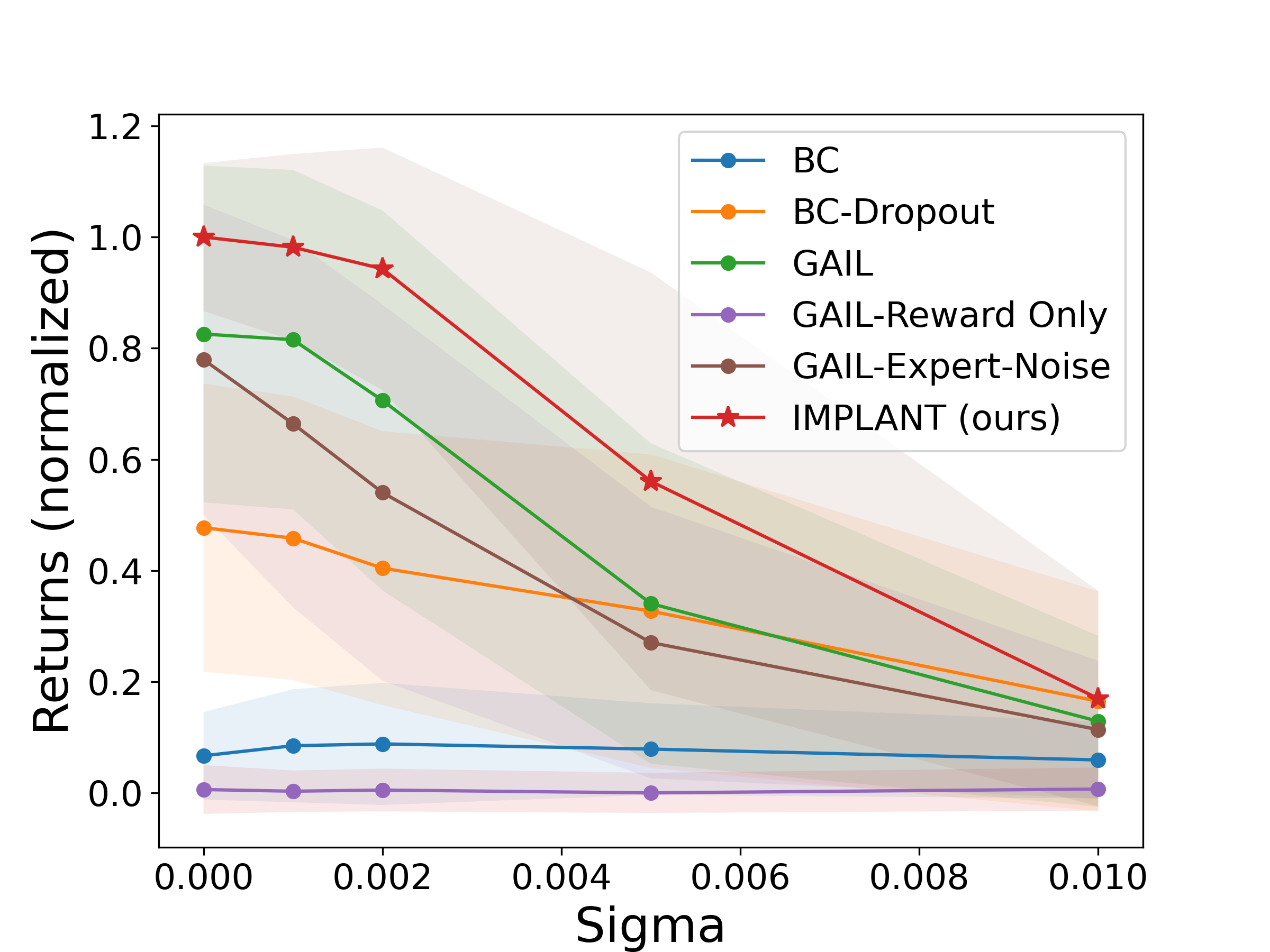}
    \caption{Walker2d: Transition Noise}
    \end{subfigure}
    \caption{Avg. return of imitation learning algorithms on motor and transition noise; shaded region shows std. error over 5 runs.
    % Comparison of imitation learning algorithms on zero-shot generalization with perturbations in test environments
    }
    \label{fig:1}
\end{figure*}

\subsection{Zero-Shot Transfer:  Motor and Transition Noise}~\label{exp:noise}
Next, we consider two kinds of noisy perturbations motivated by real-world applications in sim2real. 
First, we perturb the intended actions via \textbf{motor noise}~\citep{christiano2016transfer}, \eg, due to imperfect hardware, a real robot might execute a noisy version of the action proposed by the agent.
We implement this scenario by adding independent Gaussian noise to each dimension of the executed action at test-time, \ie, $\epsilon_{\text{action}} \stackrel{i.i.d.}{\sim} \gN(0, \sigma^2)$ and we vary the noise stddev $\sigma \in [0.1, 0.2, 0.5, 1.0]$.

Second, we consider \textbf{transition noise} due to an imperfect dynamics model for a simulator that may not be able to account for perturbations due to drag or friction. Hence,  we specify the test-time dynamics to be a perturbed noisy version of the training dynamics. 
Similar to motor noise, we sample the transition noise as  $\epsilon_{\text{transition}} \stackrel{i.i.d.}{\sim} \gN(0, \sigma^2)$ with $\sigma \in [0.001, 0.002, 0.005, 0.01]$.

For ease of visualization, we show the normalized performance of the different algorithms in Figure~\ref{fig:1}. 
See Appendix \ref{ap:2} for raw absolute results.
We also include another competitive baseline ``GAIL-Expert-Noise" relevant to this scenario that artificially adds independent noise to the demonstration data for every gradient update during GAIL training.
For a very high noise level, any algorithm will naturally deteriorate in performance due to significant shift in training and testing environments.
More importantly, for modest noise levels, we find that IMPLANT outperforms the baselines in almost all cases, highlighting its robustness.

\begin{figure}[h!]
    \centering
    \centerline{\includegraphics[width=0.85\columnwidth]{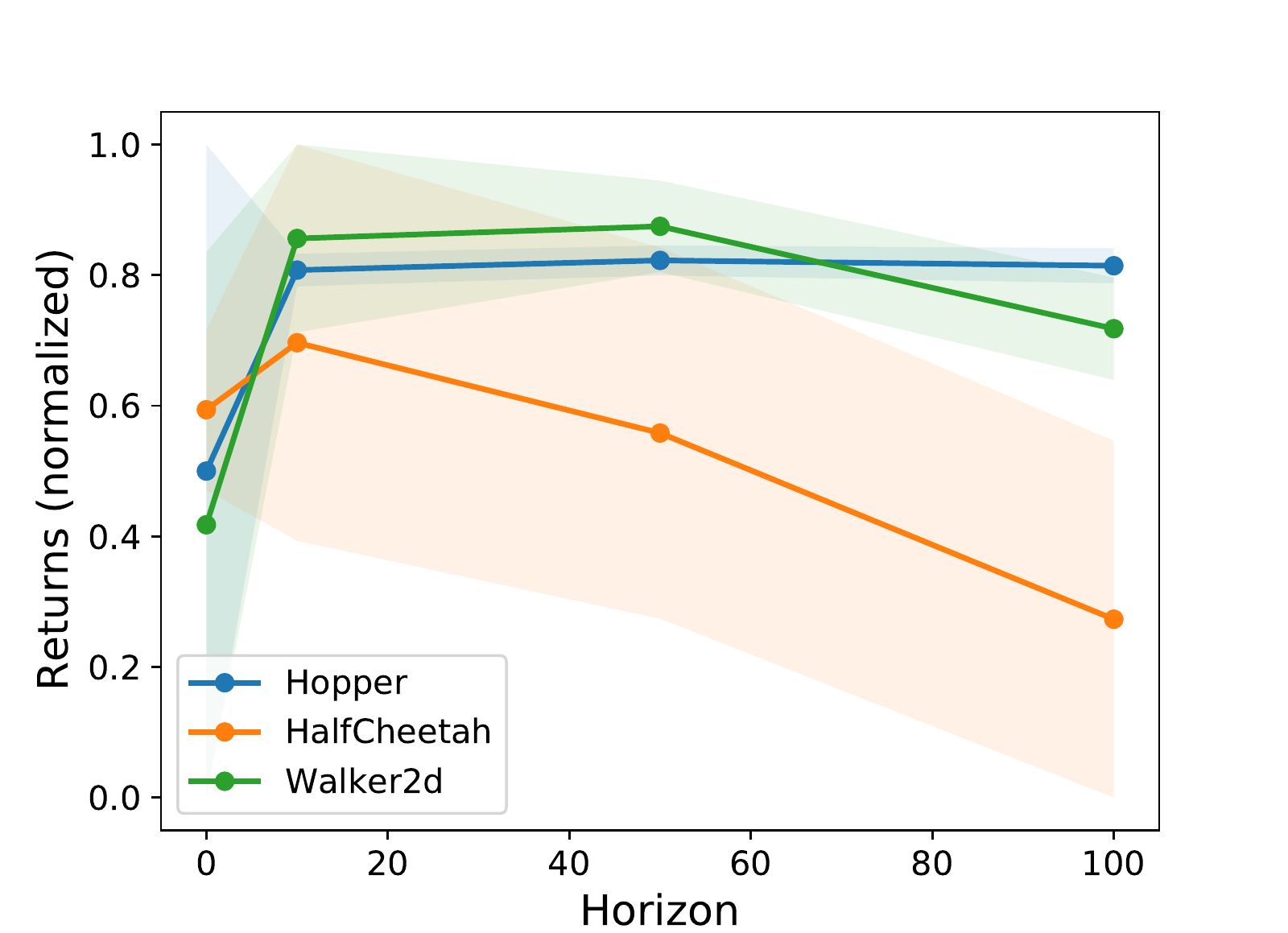}}
    \caption{Effect of planning horizon $H$ on IMPLANT; returns are normalized; shaded region shows std. error over 5 runs.
    % performance.
    }
    \label{fig:3}
    \vspace{-0.15in}
\end{figure}

\subsection{Effect of Planning Horizon}
\label{exp:design}

Finally, we analyze the effect of planning horizon on IMPLANT performance in the same setup as Section~\ref{sec:exp_default}.
Specifically, we vary the planning horizon $H \in [0, 10, 50, 100]$ for a rollout budget $B=10$.
The normalized performance curves are shown in Figure~\ref{fig:3}.
When the planning horizon is $0$, we only rely on the terminal value function for estimating returns.
Conversely, for large planning horizons (\eg, $H=100$), the returns are dominated by rewards accumulated at every time step.
We observe that picking neither a very large horizon $(h\geq100)$ nor a very small one $(h=0)$ results in optimal performance, suggesting imperfections in both the learned reward and value functions and the sweet-spot for the planning horizon is typically between the extremes.

% Finally, we will examine how the hyperparameters of our algorithm affect its performance, and we will focus the ablation study on Hopper. HalfCheetah and Walker2d hold similar trends to Hopper, so we will include the results of them in Appendix \ref{ap:2}. We study the variations on three hyperparameters of Algorithm \ref{alg:implant}: rollout policy $\pitheta \in [GAIL, random]$, rollout budget $B \in [2, 10, 20]$, and horizon $H \in [0, 10, 50, 100]$. Figure \ref{fig:2} shows that $B = 20$ and $H = 50$ works best in Hopper, and that planning with a GAIL policy outperforms planning with a random policy with the same configurations. 
 
% This brings up an interesting point: neither the policy or the value and reward function alone can perform as well as IMPLANT. Especially in transfer settings, IMPLANT exhibits outstanding performance by combining the two parts. To further demonstrate this insight, we show the results of running GAIL (policy only), IMPLANT with random policy (reward and value function only), and IMPLANT on the ``causal confusion'' setting from Section \ref{exp:causal}. Both the rendered frames and the plot of return over time in Figure \ref{fig:3} signifies that IMPLANT as a whole outperforms either of its ``sub-part''.

\section{Discussion \& Related Work}~\label{sec:related}
% \vspace{-0.25in}
% imitation advancements: airl, google latest 
% robustness in imitation general
% irl downstream cases: anca's work
% planning algs: loop
% lerrel papers
% Traditionally, algorithms for imitation learning are either completely or partially model-free during train, as in behavioral cloning~\citep{pomerleau1991efficient}, or make use inferred reward and dynamic models only during training, as in inverse reinforcement learning~\citep{ng2000algorithms,abbeel2004apprenticeship,ratliff2006maximum}.
% The former is prone to distribution shifts, whereas the latter is not able to account 
% Our work introduces a novel model-based perspective to imitation learning. 
% The use of the `model' terminology here alludes to the use of the reward function and transition dynamics during both training and execution.
% Borrowing the terminology from \citet{sutton2018reinforcement}, these phases are also referred to as \textit{background planning} and \textit{decision-time planning}.
Traditionally, algorithms for imitation learning fall into one of two categories.
They are either completely model-free during both training and execution, e.g., behavioral cloning and its variants~\citep{pomerleau1991efficient,ross2011reduction}.
Alternatively, they are model-based in the sense that they utilize dynamics and (inferred) rewards models during training, but are model-free during execution, as in inverse reinforcement learning~\citep{ng2000algorithms,ratliff2006maximum,ziebart2008maximum}.
Our work introduces a novel model-based perspective to imitation learning where the reward and transition models are used \textit{both} during training and execution. 
Borrowing the terminology from \citet{sutton2018reinforcement}, the use of such models during training and execution are also referred to as \textit{background} and \textit{decision-time} planning respectively.

While imitation via background planning has been used for control in complex environments~\citep{abbeel2004apprenticeship,ratliff2009learning,ho2016generative,choudhury2018data}, we showed that decision-time planning in IMPLANT can further improve the data efficiency and robustness of the learned policies.
There have also been several alternate attempts for characterizing and enhancing the robustness of imitation policies. 
For example, \citet{fu2017learning} seek robustness in the sense of recovering the true reward function via adversarial imitation learning and transfer the inferred reward function to external dynamics in the non-zero shot setting.
A significant body of work also considers IRL approaches that can capture the uncertainty in the reward function for safe deployment~\citep{zheng2014robust,brown2018risk,huang2018learning,lacotte2019risk,brown2020bayesian}.
While these utility-based notions are distinct from ours, they are complementary approaches to robustness that could be combined with IMPLANT in future work.
Finally, we note there has been prior work in using imitation learning \textit{for} planning in autonomous driving domain~\citep{Rhinehart2020Deep, rhinehart2021contingencies, dashora2021hybrid}. These approaches learn a dynamics model of future states given the past from the expert demonstrations and leverage this model for planning. Although these methods demonstrate their robustness in out-of-distribution environments, they require the access of a controller to output the low-level actions, which we do not assume to be available.

% \paragraph{Robust Imitation Learning.}

% \paragraph{Hierarchical }

Given the synergies between generative modeling and imitation learning as exemplified in GAIL~\citep{ho2016generative}, improvements in the former often translate into improved imitation, \eg, the use of autoencoder embeddings to improve diversity~\citep{wang2017robust}, better loss functions and architectures for stable GAN/GAIL training~\citep{pfau2016connecting,kuefler2017imitating,li2017infogail}, etc.
These modifications are conceptually complementary to the key contribution of IMPLANT to incorporate decision-time planning and are likely to further boost our performance.
In fact, decision-time planning in IMPLANT can be viewed as filtering of trajectories sampled from the policy network.
This is similar to recent work in using importance weighting for improving sample quality of a generative model~\citep{grover2017boosted,azadi2018discriminator,grover2019bias}.
However, our solution is tailored towards sequential decision making, provides flexibility in model rollouts, and deterministically picks the best outcome in line with model predictive control, unlike importance weighting filters.

% our work shares some similarities with the use of importance weighting strategies for improving sample quality of a generative model. 
% The importance weights are learned via the use of binary classifiers for density ratio estimation.

% However, our planning 

% seen as a form of 
% These techniques have shown to be effective for improving sample quality 
\section{Conclusion}

We presented \textit{Imitation with Planning at Test-time} (IMPLANT), a new meta-algorithm for imitation learning that uses decision-time planning to mitigate compounding errors of any base IRL-based imitation learning algorithm.
Unlike existing approaches, IMPLANT is truly model-based in the sense of utilizing the inferred rewards and dynamics model both during training and execution.
While decision-time planning is in general \textit{slower} than simply executing a feed-forward policy, we argue that it can be much more accurate and robust than model-free execution of imitation policies.
We demonstrated that IMPLANT matches or outperforms existing benchmark imitation learning algorithms with very few expert trajectories.
Finally, we empirically demonstrated the robustness of IMPLANT via its impressive performance at zero-shot generalization in several challenging perturbation settings involving causal confusion~\citep{de2019causal} and noisy perturbations to the environment dynamics and policy execution.
% agents learned via IMPLANT outperform existing approaches 
% but also stands out in zero-shot generalization with perturbations in test-time dynamics in various control environments. Moreover, we pointed out that IMPLANT is an incredibly compatible algorithm and has great potentials in inheriting advantages from any IL algorithm.
% Finally, we introduced a brand new model-based perspective to imitation learning. IMPLANT, being one of the first attempts to utilize the learned reward and value functions at decision-time, will hopefully inspire new approaches that further push the boundaries of performance and robustness of an imitation learning agent.  

\newpage
\bibliography{arxiv_main}

\newpage
\appendices
\section{Additional Experimental Details}\label{ap:1}
\begin{table}[ht]
    \centering
    \caption{Environment characteristics.}
    \vspace{0.5em}
    \begin{tabular}{lccc}
    \toprule

    Environment & Observation space & Action space \\
    \midrule
    Hopper & Box(11, ) & Box(3, ) \\
    HalfCheetah & Box(17, ) & Box(6, ) \\
    Walker2d & Box(17, ) & Box(6, ) \\
    \bottomrule
    \end{tabular}
    \label{tab:3}
\end{table}
\subsection{Environments and Expert Data}

As mentioned above, we consider 3 continuous tasks from OpenAI Gym \citep{brockman2016openai} simulated with MuJoCo \citep{todorov2012mujoco}: Hopper, HalfCheetah, and Walker2d. We acquire expert data by training a SAC \citep{haarnoja2018soft} agent with ground truth reward and then recording its rollouts. Table \ref{tab:3} lists more detailed information about each environment, and Table \ref{tab:4} contains information about the expert demonstrations we use for training all of the agents.

\begin{table*}[h!]
    \centering
    \caption{Expert dataset characteristics.}
    \begin{tabular}{lcccc}
    \toprule

    Environment & \# trajectories & \# state-action pairs & expert performance \\
    \midrule
    Hopper & $4$ & $200$ & $3570$\\
    HalfCheetah & $4$ & $200$ & $9892$\\
    Walker2d & $4$ & $200$ & $4585$\\
    
    \bottomrule
    \end{tabular}
    \label{tab:4}
\end{table*}

\subsection{Hyperparameters and Network Architectures}
We use a 2-layer MLP with \emph{tanh} activations and 100 hidden units for all of our policy networks. For BC, we use a learning rate of $10^{-4}$ across all environments. We use a dropout rate of 0.2 for our BC-Dropout agent. The hyperparameters of GAIL are listed in Table \ref{tab:5}. 

For our IMPLANT agent, we directly utilize the value function, reward function, and policy from a trained GAIL agent. 
Further, we set $B = 20$ and $H = 50$ for Hopper, $B=2$ and $H=10$ for HalfCheetah, and  $B=20$ and $H=50$ for Walker2d for MPC planning. 

\newpage
\section{Additional Results}\label{ap:2}

\subsection{Raw Performance of Motor Noise and Transition Noise}
The complete results for Section \ref{exp:noise} can be found in Table \ref{tab:6}, \ref{tab:7}, \ref{tab:8}.
\begin{figure}[th]
    \centering
    \centerline{\includegraphics[width=0.85\columnwidth]{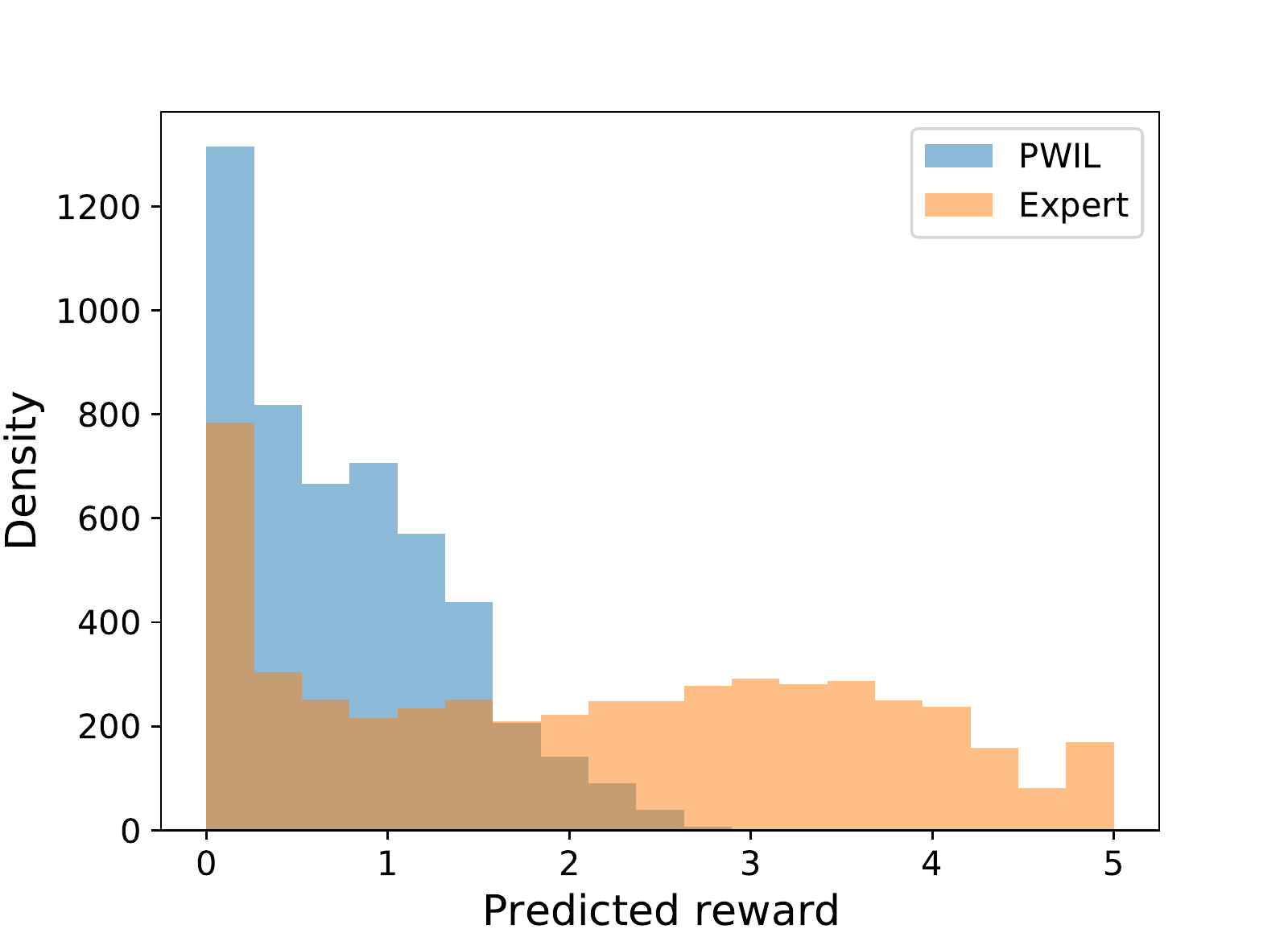}}
    \caption{Distribution of predicted rewards for trained PWIL policy vs. expert in Hopper.
    % performance.
    }
    \label{fig:4}
    \vspace{-0.3in}
\end{figure}
\subsection{Robustifying other IRL algorithms}
% To demonstrate that IMPLANT is a meta-algorithm for imitation learning and works well with a variety of imitation learning algorithms, 
In principle, IMPLANT is a meta-algorithm and hence can be used as a wrapper for improving the robustness of any IRL-based imitation learning algorithm.
To demonstrate the general-purposeness, we perform planning with a very recent state-of-the-art imitation learning algorithm, Primal Wasserstein Imitation Learning (PWIL)~\citep{dadashi2020primal} in Hopper environment. Unlike GAIL, PWIL is a non-adversarial algorithm and exhibits impressive sample efficiency in terms of interactions with the environment.

Figure~\ref{fig:4} shows a mismatch in the distributions of rewards for policy rollouts and expert, as predicted by the reward function for PWIL, suggesting the benefits of planning using the same reward function at test-time. 
Tables~\ref{tab:pwilacts} and \ref{tab:pwiltrans} show the improvements in performance via IMPLANT for the setups with motor noise and transition noise perturbations respectively.
% Figure~\ref{fig:pwilvsimplant} shows the improvements in performance via IMPLANT for the setups with motor noise and transition noise perturbations.
% Raw results for above setups are shown in Table~\ref{tab:pwilacts} and \ref{tab:pwiltrans}, respectively.
% \vspace{-0.15in}

\begin{table*}[h!]
    \centering
    \caption{Hyperparameters used for GAIL training.}
    \begin{tabular}{lcccc}
    \toprule
    Parameters & Hopper & HalfCheetah & Walker2d \\
    \midrule
    Discriminator network & 100-100 MLP & 100-100 MLP & 100-100 MLP \\
    Discriminator entropy coeff.& 0.01 & 0.01 & 0.01 \\
    Batch size  &1024 & 50000 & 50000\\
    Max kl & 0.01& 0.01&0.01 \\
    CG steps/damping & 10, 0.01 & 10, 0.1 & 10, 0.1\\
    Entropy coeff. & 0.0 & 0.0 & 0.0 \\
    Value fn. steps/step size & 3, 3e-4 & 5,3e-4 & 5,3e-4\\
    Generator steps & 3 & 3 & 3 \\
    Discriminator steps & 1 & 1 & 1 \\
    $\lambda$ & 0.98 & 0.97 & 0.97 \\
    $\gamma$ & 0.99 & 0.995 & 0.995 \\

    \bottomrule
    \end{tabular}
    \label{tab:5}
\end{table*}

% \begin{figure}[t]
%     \centering
%     \centerline{\includegraphics[width=0.85\columnwidth]{figs/implantvspwil.pdf}}
%     \caption{Average returns of PWIL vs IMPLANT on motor noise in Hopper environment. Shaded region shows standard error over 5 runs.
%     % performance.
%     }
%     \label{fig:5}
%     \vspace{-0.15in}
% \end{figure}

% \begin{figure}[t]
%     \centering
%     \centerline{\includegraphics[width=0.85\columnwidth]{figs/implantvspwildynamics.pdf}}
%     \caption{Average returns of PWIL vs IMPLANT on transition noise in Hopper environment. Shaded region shows standard error over 5 runs.
%     % performance.
%     }
%     \label{fig:6}
%     \vspace{-0.15in}
% \end{figure}

% \begin{figure}[h!]
%     \centering
%     \begin{subfigure}[b]{.8\linewidth}
%         \centering
%         \begin{subfigure}[b]{\linewidth}
%         \centering
%         \includegraphics[width=\linewidth]{figs/implantvspwil.pdf}
%         \caption{Hopper: Motor Noise} \label{fig:pwilmotor}
%         \end{subfigure}
%         \vspace{-0.15in}
%         \centering
%         \begin{subfigure}[b]{\linewidth}
%         \centering
%         \includegraphics[width=\linewidth]{figs/implantvspwildynamics.pdf}
%         \caption{Hopper: Transition Noise} \label{fig:pwiltransition}
%         \end{subfigure}
%     \end{subfigure}
%     \caption{Average returns of PWIL vs IMPLANT on motor noise (left) and transition noise(right) in Hopper environment. Shaded region shows standard error over 5 runs.}\label{fig:pwilvsimplant}
%     \vspace{-0.15in}
% \end{figure}

\begin{table*}[h]
%  \vspace{-15mm}
    \centering
        \caption{Raw results of Figure \ref{fig:1} in Hopper environment}
    % \vspace{-4mm}
    \subfloat[Hopper with motor noise]{
    \begin{tabular}{lccccc}
    \toprule
    Sigma & 0.0 & 0.1 & 0.2 & 0.5 & 1.0 \\
    \midrule
    BC                  & $127 \pm 85 $ & $114 \pm 64$ & $117 \pm 61$ & $179 \pm 105$ & $123 \pm 82$ \\
    BC-Dropout          & $169 \pm 105$ & $156 \pm 85$ & $163 \pm 95$ & $ 158 \pm 89$ & $142 \pm 94$ \\
    GAIL                & $3506 \pm 337$& $2572 \pm 1008$ & $1360 \pm 687$ & $598 \pm 243$ & $252 \pm 128$ \\
    % GAIL-Dropout        & $127 \pm 50$  & $128 \pm 49$ & $128 \pm 46$ & $127 \pm 53$ & $93 \pm 69$ \\ 319, 123	293, 87	268, 92	160, 113	49, 54
    GAIL-Reward Only & $319 \pm 123$ &	$293 \pm 87$ &  $268 \pm 92$ &	$160 \pm 113$ &	$49 \pm 54$ \\
    GAIL-Expert-Noise   & $3602 \pm 46$ & $2716 \pm 921$ & $1449 \pm 667$ & $\textbf{618} \pm 264$ & $218 \pm 136$ \\
    IMPLANT (ours)      & $\textbf{3633} \pm 50$ & $\textbf{3209} \pm 714$ & $\textbf{1764} \pm 856$ & $596 \pm 234$ & $\textbf{269} \pm 130$ \\
    
    \bottomrule
    \end{tabular}
    }
    
  \vspace{2mm}
  \centering
  \subfloat[Hopper with transition noise]{
  \begin{tabular}{lccccc}
    \toprule
    Sigma & 0.0 & 0.001 & 0.002 & 0.005 & 0.01 \\
    \midrule
    BC                  & $127 \pm 85 $ & $123 \pm 81$ & $116 \pm 65$ & $137 \pm 98$ & $114 \pm 84$ \\
    BC-Dropout          & $169 \pm 105$ & $175 \pm 113$ & $165 \pm 99$ & $ 161 \pm 98$ & $151 \pm 101$ \\
    GAIL                & $3506 \pm 337$& $3209 \pm 756$ & $2672 \pm 1012$ & $\textbf{1377} \pm 901$ & $\textbf{616} \pm 563$ \\
    % GAIL-Dropout        & $127 \pm 50$  & $128 \pm 50$ & $128 \pm 49$ & $131 \pm 52$ & $129 \pm 66$ 319, 123	884, 230	804, 269	655, 296	349, 215
    GAIL-Reward Only & $319 \pm 123$ &	$884 \pm 230$ &  $804 \pm 269$ &	$665 \pm 296$ &	$349 \pm 215$ \\
    GAIL-Expert-Noise   & $3576 \pm 36$ & $2966 \pm 1089$ & $2160 \pm 1311$ & $875 \pm 865$ & $323 \pm 357$ \\
    IMPLANT (ours)      & $\textbf{3633} \pm 50$ & $\textbf{3557} \pm 313$ & $\textbf{2844} \pm 915$ & $1301 \pm 810$ & $598 \pm 467$ \\
    
    \bottomrule
    \end{tabular}
    }
        \label{tab:6}
\end{table*}

\begin{table*}[h!]
    \centering
        \caption{Raw results of Figure \ref{fig:1} in HalfCheetah environment}
%     %Cheetah
% Sigma	0	0.1	0.2	0.5	1
% BC	-359, 247	-375, 238	-355, 241	-409, 175	-873, 102
% BC Dropout	-99, 229	-114, 232	-145, 252	-272, 196	-811, 93
% GAIL	4059, 728	3722, 650	2960, 420	1134, 270	-413, 226
% GAIL w/ expert noise	3481, 635	3314, 500	2806, 405	1190, 262	-411, 193
% IMPLANT w/ random		-284, 77 -269, 111	-293, 106	-399, 97	-833, 77
% IMPLANT	5240, 924	4317, 486	3259, 382	1182, 237	-416, 217
    \vspace{2mm}
    \centering
    \subfloat[HalfCheetah with motor noise]{
    \begin{tabular}{lccccc}
    \toprule
    Sigma & 0.0 & 0.1 & 0.2 & 0.5 & 1.0 \\
    \midrule
    BC                  & $-359 \pm 247 $ & $-375 \pm 238$ & $-355 \pm 241$ & $-409 \pm 175$ & $-873 \pm 102$ \\
    BC-Dropout          & $-99 \pm 229$ & $-114 \pm 232$ & $-145 \pm 252$ & $ -272 \pm 196$ & $-811 \pm 193$ \\
    GAIL                & $4059 \pm 728$& $3722 \pm 650$ & $2960 \pm 420$ & $1134 \pm 270$ & $-413 \pm 226$ \\
    % GAIL-Dropout        & $-1 \pm 1$  & $-4 \pm 9$ & $-29 \pm 24$ & $-226 \pm 74$ & $\textbf{-648} \pm 72$ \\
    GAIL-Reward Only & $-284\pm 77$	& $-269\pm 111$ & 	$-293 \pm 106$ &          	$-399\pm 97$ &  	$-833\pm 77$ \\
    GAIL-Expert-Noise   & $3481 \pm 635$ & $3314 \pm 500$ & $2806 \pm 405$ & $\textbf{1190} \pm 262$ & $\textbf{-411}\pm 193$ \\
    IMPLANT (ours)      & $\textbf{5240} \pm 924$ & $\textbf{4317} \pm 486$ & $\textbf{3259} \pm 382$ & $1182 \pm 237$ & $-416 \pm 217$ \\
    \bottomrule
    \end{tabular}
    }
    
 \vspace{4mm}
 \centering
  \subfloat[HalfCheetah with transition noise]{
  \begin{tabular}{lccccc}
    \toprule
%     Sigma	0	0.001	0.002	0.005	0.01
% BC	-359, 247	-368, 244	-360, 271	-402, 214	-426, 208
% BC Dropout	-99, 229	-138, 209	-115, 220	-128, 219	-139, 192
% GAIL	4059, 728	4113, 600	4066, 748	3865, 687	3506, 640
% GAIL w/ expert noise	3481, 635	3527, 527	3429, 616	3330, 633	3064, 683
% IMPLANT w/ random	-284, 77	-271, 107	-266, 109	-287, 107	-293, 117
% IMPLANT	5240, 924	5212, 544	5080, 486	4716, 498	4087, 444
    Sigma & 0.0 & 0.001 & 0.002 & 0.005 & 0.01 \\
    \midrule
    BC                  & $-359 \pm 247 $ & $-368 \pm 244$ & $-360 \pm 271$ & $-402 \pm 214$ & $-426 \pm 208$ \\
    BC-Dropout          & $-99 \pm 229$ & $-138 \pm 209$ & $-115 \pm 220$ & $ -128 \pm 219$ & $-139 \pm 192$ \\
    GAIL                & $4059 \pm 728$& $4113 \pm 600$ & $4066 \pm 748$ & $3865 \pm 687$ & $3506 \pm 640$ \\
    % GAIL-Dropout        & $-1 \pm 1$  & $-4 \pm 9$ & $-29 \pm 24$ & $-226 \pm 74$ & $\textbf{-648} \pm 72$ \\
    GAIL-Reward Only & $-284\pm 77$	& $-271\pm 107$ & 	$-266 \pm 109$ &          	$-287\pm 107$ &  	$-293\pm 117$ \\
    GAIL-Expert-Noise   & $3481 \pm 635$ & $3527 \pm 527$ & $3429 \pm 616$ & $3330 \pm 633$ & $3064 \pm 683$ \\
    IMPLANT (ours)      & $\textbf{5240} \pm 924$ & $\textbf{5212} \pm 544$ & $\textbf{5080} \pm 486$ & $\textbf{4716} \pm 498$ & $\textbf{4087} \pm 444$ \\
    \bottomrule
    \end{tabular}
    }

     \label{tab:7}
\end{table*}
   \vspace{50mm}
\begin{table*}[h!]
%     % Walker
    \centering
        \caption{Raw results of Figure \ref{fig:1} in Walker2d environment}
    \centering
    \subfloat[Walker2d with motor noise]{
    \begin{tabular}{lccccc}
    \toprule
% BC	153, 162	152, 163	144, 161	55, 76	10, 13
% BC Dropout	344, 72	350, 80	345, 133	211, 139	25, 37
% GAIL	3847, 635	3603, 830	2626, 1214	608, 300	185, 155
% GAIL w/ expert noise	3437, 787	3456, 822	2570, 1285	572, 328	171, 166
% IMPLANT w/ random	2, 5	32, 94	25, 78	12, 33	3, 9
% IMPLANT	4403, 242	3666, 991	2816, 1019	596, 336	214, 148
    Sigma & 0.0 & 0.1 & 0.2 & 0.5 & 1.0 \\
    \midrule
    BC	& $153\pm 162$ & $152\pm 163$ & $144 \pm 161$ & $55\pm 76$ & $10\pm 13$\\
    BC-Dropout	& $344\pm 72$ & $350\pm 80$ & $345\pm 133$ & $211\pm 139$	 & $25\pm 37$\\
    GAIL	& $3847\pm 635$ & $3603\pm 830$ & $2626\pm 1214	$ & $\textbf{608}\pm 300	$ & $185\pm 155$\\
    % GAIL-Dropout	& $272\pm 16	$&$271\pm 10	$&$277\pm 24	$&$264\pm 67	$&$241\pm 103$\\
    GAIL-Reward Only & $2 \pm 5$ &	$32 \pm 94$ &  $25 \pm 78$ &	$12 \pm 33$ &	$3 \pm 9$ \\
    GAIL-Expert-Noise	& $3437\pm 787$ & $3456\pm 822$ & $2570\pm 1285$ & $572\pm 328$ & $171\pm 166$ \\

    IMPLANT (ours)	& $\textbf{4403}\pm 242$ & $\textbf{3666}\pm 991$ & $\textbf{2816}\pm 1019$ & $596\pm 336$ & $\textbf{214}\pm 148$ \\
    \bottomrule
    \end{tabular}
    }

% Dynamics Noise					
% Sigma	0	0.001	0.002	0.005	0.01
% BC	153, 162	162, 171	156, 167	145, 166	119, 153
% BC Dropout	344, 72	338, 102	327, 124	302, 175	261, 168
% GAIL	3847, 635	3835, 671	3689, 921	2096, 1412	856, 938
% GAIL w/ expert noise	3437, 787	3422, 779	3317, 879	2387, 1272	1147, 876
% IMPLANT w/ random	2, 5	18, 67	22, 79	41, 95	18, 67
% IMPLANT	4403, 242	4373, 307	4202, 671	2653, 1337	1293, 1098
    \vspace{4mm}
    \centering
    \subfloat[Walker2d with transition noise]{
  \begin{tabular}{lccccc}
    \toprule
    Sigma & 0.0 & 0.001 & 0.002 & 0.005 & 0.01 \\
    \midrule
    BC	& $153\pm 162$ & $162\pm 171$ & $156 \pm 167$ & $145\pm 166$ & $119\pm 153$\\
    BC-Dropout	& $344\pm 72$ & $338\pm 102$ & $327\pm 124$ & $302\pm 175$	 & $261\pm 168$\\
    GAIL	& $3847\pm 635$ & $3835\pm 671$ & $3689\pm 921	$ & $2096\pm 1412	$ & $856\pm 938$\\
    % GAIL-Dropout	& $272\pm 16	$&$271\pm 10	$&$277\pm 24	$&$264\pm 67	$&$241\pm 103$\\
    GAIL-Reward Only & $2 \pm 5$ &	$18 \pm 67$ &  $22 \pm 79$ &	$41 \pm 95$ &	$18 \pm 67$ \\
    GAIL-Expert-Noise	& $3437\pm 787$ & $3422\pm 779$ & $3317\pm 879$ & $2387\pm 1272$ & $1147\pm 876$ \\

    IMPLANT (ours)	& $\textbf{4403}\pm 242$ & $\textbf{4373}\pm 307$ & $\textbf{4202}\pm 671$ & $\textbf{2653}\pm 1337$ & $\textbf{1293}\pm 1098$ \\
    \bottomrule
    \end{tabular}
   }
    \vspace{4mm}
    \label{tab:8}
\end{table*}

\begin{table*}[ht]
    \centering
        \caption{Imitation via PWIL on Hopper with motor noise}
    % \vspace{-4mm}
    % \subfloat[Hopper with motor noise]{
    \begin{tabular}{lccccc}
    \toprule
    Sigma & 0.0 & 0.1 & 0.2 & 0.5 & 1.0 \\
    \midrule
    PWIL & $3552 \pm 200$ &	$2189 \pm 1236$ &  $935 \pm 706$ &	$\textbf{393} \pm 119$ &	$218 \pm 86$ \\
    IMPLANT w/ PWIL (ours)      & $\textbf{3563} \pm 87$ & $\textbf{2637} \pm 1140$ & $\textbf{1110} \pm 817$ & $390 \pm 103$ & $\textbf{232} \pm 82$ \\
    \bottomrule
    \end{tabular}
    % }
    \label{tab:pwilacts}
\end{table*}

\begin{table*}
%  \vspace{-15mm}
    \centering
        \caption{Imitation via PWIL on Hopper with transition noise}
    % \vspace{-4mm}
    % \subfloat[Hopper with transition noise]{
    \begin{tabular}{lccccc}
    \toprule
    Sigma & 0.0 & 0.001 & 0.002 & 0.005 & 0.01 \\
    \midrule
    PWIL & $3552 \pm 200$ &	$3011 \pm 958$ &  $1961 \pm 1144$ &	$807 \pm 500$ &	$\textbf{469} \pm 319$ \\
    IMPLANT w/ PWIL (ours)      & $\textbf{3563} \pm 87$ & $\textbf{3226} \pm 739$ & $\textbf{1982} \pm 1079$ & $\textbf{942} \pm 649$ & $452 \pm 293$ \\
    \bottomrule
    \end{tabular}
    % }
    \label{tab:pwiltrans}
\end{table*}

\end{document}